%% file: paper.tex
\documentclass[
]{article}

\usepackage[utf8]{inputenc}
\usepackage[T1]{fontenc}

\input{preamble/00_begin}
\input{preamble/01_style}
\input{preamble/02_text}

\input{preamble/03_figures}
\input{preamble/04_tables}
\input{preamble/05_algorithms}
\input{preamble/06_mathematics}
\input{preamble/07_draft}

\input{preamble/08_paper_macros}

\input{preamble/99_end}

\newcommand{\papertitle}{Computation-Aware Kalman Filtering with Model Selection for Neural Dynamics}

\probnumtitle{\papertitle}

\probnumauthors{
  JR Huml \affiliation{Department of Statistics, Columbia University; Zyphra Technologies}\\
  Jonathan Wenger \affiliation{Department of Statistics, Columbia University; Zuckerman Institute}\\
  John P. Cunningham \affiliation{Department of Statistics, Columbia University; Zuckerman Institute}
}

\probnumabstract{%
  \input{00_abstract}
}

\begin{document}

% -----------------------------------------------------------------------------
% Old \maketitle (commented out; ProbNum prints title automatically)
% -----------------------------------------------------------------------------
% \maketitle

% -----------------------------------------------------------------------------
% ICML commands (commented out; kept for reference)
% -----------------------------------------------------------------------------
% \icmltitlerunning{\papertitle}
%
% \twocolumn[
%   \icmltitle{\papertitle}
%   \icmlsetsymbol{equal}{*}
%   \begin{icmlauthorlist}
%   \icmlauthor{Jonathan Huml}{equal,columbia}
%   \icmlauthor{Jonathan Wenger}{equal,columbia,zuckerman}
%   \icmlauthor{John P. Cunningham}{columbia,zuckerman}
%   \end{icmlauthorlist}
%   \icmlaffiliation{columbia}{Department of Statistics, Columbia University}
%   \icmlaffiliation{zuckerman}{Zuckerman Institute, Columbia University}
%   \icmlcorrespondingauthor{Firstname1 Lastname1}{first1.last1@xxx.edu}
%   \icmlkeywords{Machine Learning, ICML}
%   \vskip 0.3in
% ]
% %\printAffiliationsAndNotice{}  
% %\printAffiliationsAndNotice{\icmlEqualContribution}

% -----------------------------------------------------------------------------
% Old abstract env (commented out; ProbNum renders the abstract itself)
% -----------------------------------------------------------------------------
% \begin{abstract}
%   \input{00_abstract}
% \end{abstract}

\input{01_introduction}

\input{02_background}

\input{03_model}
\input{04_experiments}

\input{05_conclusion}

\bibliographystyle{unsrtnat}
\bibliography{references}

\newpage
\onecolumn
\input{98_appendix}

\end{document}

%% file: preamble/00_begin.tex
\usepackage{ifdraft}

%% file: preamble/01_style.tex
\ifdraft{
  \usepackage{ProbNum26}
}{
  \usepackage[final]{ProbNum26}
}

%% file: preamble/02_text.tex
\usepackage{microtype}

\usepackage[inline]{enumitem}

\hypersetup{
  unicode,                    % Use unicode for links
  pdfborder       = {0 0 0},  % Suppress border around pdf
  bookmarksdepth  = subsection,
  bookmarksopen   = true,     % Expand the bookmarks as soon as the pdf file is opened
  linktoc         = all,      % Toc entries and numbers links to pages
  breaklinks      = true,
  colorlinks      = true,
  linkcolor       = linkcolor,
  citecolor       = citecolor,
  urlcolor        = urlcolor,
  }
\usepackage[numbered]{bookmark}  % Add chapter/section numbers to PDF outline

\usepackage[nottoc, section]{tocbibind}
\usepackage[numbers,sort&compress]{natbib}

%% file: preamble/03_figures.tex
\usepackage{wrapfig}

\usepackage{caption}

\usepackage[
  labelformat=simple
]{subcaption}

\usepackage{xcolor}

\definecolor{TUred}{RGB}{165,30,55}
\definecolor{TUgold}{RGB}{180,160,105}
\definecolor{TUdark}{RGB}{50,65,75}
\definecolor{TUgray}{RGB}{175,179,183}

\definecolor{TUdarkblue}{RGB}{65,90,140}
\definecolor{TUblue}{RGB}{0,105,170}
\definecolor{TUlightblue}{RGB}{80,170,200}
\definecolor{TUlightgreen}{RGB}{130,185,160}
\definecolor{TUgreen}{RGB}{125,165,75}
\definecolor{TUdarkgreen}{RGB}{50,110,30}
\definecolor{TUocre}{RGB}{200,80,60}
\definecolor{TUviolet}{RGB}{175,110,150}
\definecolor{TUmauve}{RGB}{180,160,150}
\definecolor{TUbeige}{RGB}{215,180,105}
\definecolor{TUorange}{RGB}{210,150,0}
\definecolor{TUbrown}{RGB}{145,105,70}

\definecolor{MPLblue}{HTML}{1f77b4}
\definecolor{MPLorange}{HTML}{ff7f0e}
\definecolor{MPLgreen}{HTML}{2ca02c}
\definecolor{MPLred}{HTML}{d62728}
\definecolor{MPLpurple}{HTML}{9467bd}

\definecolor{SNSblue}{rgb}{0.1216, 0.4666, 0.7059}
\definecolor{SNSorange}{rgb}{1.0, 0.4980, 0.0549}
\definecolor{SNSgreen}{rgb}{0.1725, 0.6274, 0.1725}
\definecolor{SNSred}{rgb}{0.84, 0.15, 0.16}
\definecolor{SNSpurple}{rgb}{0.58, 0.40, 0.74}

\definecolor{SNSblue_shaded}{HTML}{8ebad9}
\definecolor{SNSorange_shaded}{HTML}{ffcea3}
\definecolor{SNSgreen_shaded}{HTML}{cae7ca}
\definecolor{SNSred_shaded}{HTML}{ea9293}

\definecolor{CUblue}{RGB}{29,79,145}
\definecolor{CUlightblue}{RGB}{185,217,235}
\definecolor{CUdarkblue}{RGB}{0,48,135}
\definecolor{CUgray}{RGB}{117,120,123}
\definecolor{CUlightgray}{RGB}{208,208,206}
\definecolor{CUdarkgray}{RGB}{83,86,90}
\definecolor{CUmagenta}{RGB}{174,37,115}
\definecolor{CUyellow}{RGB}{255,152,0}
\definecolor{CUorange}{RGB}{185,71,0}
\definecolor{CUred}{RGB}{166,10,61}
\definecolor{CUgreen}{RGB}{118,136,29}

\makeatletter
\@ifclassloaded{beamer}{
    \colorlet{citecolor}{CUblue}
    \colorlet{citelightcolor}{CUlightblue}
    \colorlet{linkcolor}{CUblue}
    \colorlet{linklightcolor}{CUlightblue}
    \colorlet{urlcolor}{CUblue}
    \colorlet{urllightcolor}{CUlightblue}
}{
    \colorlet{citecolor}{MPLblue}
    \colorlet{linkcolor}{black}
    \colorlet{urlcolor}{MPLblue}
}
\makeatother

\usepackage{tikz}

\usetikzlibrary{positioning}

\usetikzlibrary{arrows,shapes,plotmarks,decorations.pathmorphing,automata,chains,fit,backgrounds,calc,positioning,fadings}

\tikzset{every picture/.style={node distance=2cm,shorten >= 1pt, shorten <= 1pt}}
\tikzset{>=stealth'}
\tikzstyle{graphnode} =
[circle,draw=TUdark,minimum size=22pt,text centered,text width=22pt,inner sep=0pt, text=TUdark]
\tikzstyle{var}   =[graphnode,fill=white]
\tikzstyle{obs}   =[graphnode,fill=TUdarkblue,text=white]
\tikzstyle{act}   =[rectangle,draw=TUdark,text=white,minimum
size=22pt,text centered, text width=22pt,inner sep=0pt]
\tikzstyle{fac}   =[rectangle,draw=TUdark,fill=TUgreen,minimum size=5pt]
\tikzstyle{facprior} =[rectangle,draw=TUdark,fill=TUdark,text=white,minimum size=5pt]
\tikzstyle{edge}  =[draw=TUdark,-]
\tikzstyle{prior} =[rectangle, draw=TUdark, fill=TUdark, minimum size=
5pt, inner sep=0pt]
\tikzstyle{dirprior} = [circle, draw=TUdark, fill=TUdark, minimum
size=5pt, inner sep=0pt]

\tikzfading[name=fade top,bottom color=transparent!0,top color=transparent!75]

%% file: preamble/04_tables.tex
\usepackage{siunitx}
\usepackage{booktabs}
\usepackage{multirow}
\usepackage{titletoc}		% TOC for appendix

%% file: preamble/05_algorithms.tex
\usepackage{algorithm}
\newcommand{\just}[2]{\phantom{#1}\mathllap{#2}}

%% file: preamble/06_mathematics.tex
\usepackage{amsmath}
\usepackage{mathtools}

\usepackage{amssymb}
\usepackage{bm}
\usepackage{nicefrac}

\usepackage{amsthm}
\usepackage{thmtools}
\usepackage{thm-restate}

\usepackage{etoolbox}
\usepackage{xparse}

\allowdisplaybreaks    % Allow page breaks in `align` etc.

\numberwithin{equation}{section}  % Equation numbers are prefixed by the section number

\DeclarePairedDelimiter{\ps}{(}{)}

\newcommand*{\defeq}{\coloneqq}  % equal by definition ":="
\newcommand*{\rdefeq}{\eqqcolon}  % equal by definition (reversed) "=:"

\DeclarePairedDelimiterX{\set}[1]\{\}{%

#1}

\newcommand*{\setsym}[1]{\ensuremath{{\mathbb{#1}}}}

\newcommand*{\R}{\setsym{R}}

\newcommand*{\inv}{^{-1}}

\newcommand*{\closure}[1]{\overline{#1}}

\renewcommand*{\vec}[1]{{\bm{#1}}}

\newcommand*{\vb}{\vec{b}}

\newcommand*{\vm}{\vec{m}}

\newcommand*{\vr}{\vec{r}}
\newcommand*{\vs}{\vec{s}}

\newcommand*{\vw}{\vec{w}}

\newcommand*{\vy}{\vec{y}}

\newcommand*{\vmu}{\vec{\mu}}

\NewDocumentCommand{\range}{s m}{%
  \operatorname{ran}
  \IfBooleanTF{#1}{\left(}{(}
  #2
  \IfBooleanTF{#1}{\right)}{)}
}

\NewDocumentCommand{\kernel}{s m}{%
  \operatorname{ker}
  \IfBooleanTF{#1}{\left(}{(}
  #2
  \IfBooleanTF{#1}{\right)}{)}
}

\newcommand*{\mat}[1]{{\bm{#1}}}

\newcommand*{\mA}{\mat{A}}

\newcommand*{\mC}{\mat{C}}
\newcommand*{\mD}{\mat{D}}

\newcommand*{\mF}{\mat{F}}
\newcommand*{\mG}{\mat{G}}
\newcommand*{\mH}{\mat{H}}
\newcommand*{\mI}{\mat{I}}

\newcommand*{\mK}{\mat{K}}
\newcommand*{\mL}{\mat{L}}
\newcommand*{\mM}{\mat{M}}

\newcommand*{\mP}{\mat{P}}
\newcommand*{\mQ}{\mat{Q}}

\newcommand*{\mS}{\mat{S}}

\newcommand*{\mU}{\mat{U}}
\newcommand*{\mV}{\mat{V}}
\newcommand*{\mW}{\mat{W}}

\newcommand*{\mLambda}{\mat{\Lambda}}

\newcommand*{\mSigma}{\mat{\Sigma}}

\makeatletter
\DeclarePairedDelimiterXPP{\@norm}[2]{}{\lVert}{\rVert}{\ifstrempty{#2}{}{_{#2}}}{#1}
\NewDocumentCommand{\norm}{s O{} m O{}}{%
  \IfBooleanTF{#1}{%
    \@norm*{#3}{#4}%
  }{%
    \@norm[#2]{#3}{#4}%
  }%
}
\makeatother

\newcommand*{\dualop}{'}

\makeatletter
\DeclarePairedDelimiterXPP{\@inprod}[3]{}{\langle}{\rangle}{\ifstrempty{#3}{}{_{#3}}}{#1, #2}
\NewDocumentCommand{\inprod}{s O{} m m O{}}{%
  \IfBooleanTF{#1}{%
    \@inprod*{#3}{#4}{#5}%
  }{%
    \@inprod[#2]{#3}{#4}{#5}%
  }%
}
\makeatother

\newcommand*{\T}{^\top}

\newcommand*{\pinv}{^\dagger}

\NewDocumentCommand{\cfns}{m o}{%
  C (#1\IfValueT{#2}{, #2})
}

\NewDocumentCommand{\cdfns}{O{0} m o}{%
  C^{#1} (#2\IfValueT{#3}{, #3})
}

\NewDocumentCommand{\holdersp}{m o m}{%
  C^{#1\IfValueT{#2}{, #2}}(\closure{#3})
}

\RenewDocumentCommand{\L}{m o}{%
  \ensuremath{
    L_{#1}
    \IfNoValueF{#2}{\left( #2 \right)}
  }
}

\NewDocumentCommand{\sobolev}{o m o}{%
  \IfNoValueTF{#1}{
    H^{#2}
  }{
    W^{#1,#2}
  }
  \IfNoValueF{#3}{\left( #3 \right)}
}

\NewDocumentCommand{\sobolevtest}{o m o}{%
  \sobolev[#1]{#2}_0
  \IfNoValueF{#3}{\left( #3 \right)}
}

\newcommand*{\linop}[1]{\mathcal{#1}}

\newcommand*{\linfctls}[1]{{\bm{\linop{#1}}}}

\makeatletter
\DeclarePairedDelimiter{\@evallinop@oparg}{[}{]}
\DeclarePairedDelimiter{\@evallinop@fnarg}{(}{)}
\NewDocumentCommand{\evallinop}{m s O{} m s O{} d()}{%
  #1%
  \IfBooleanTF{#2}{%
    \@evallinop@oparg*{#4}%
  }{%
    \@evallinop@oparg[#3]{#4}%
  }%
  \IfValueT{#7}{%
    \IfBooleanTF{#5}{%
      \@evallinop@fnarg*{#7}%
    }{%
      \@evallinop@fnarg[#6]{#7}%
    }%
  }%
}
\makeatother

\NewDocumentCommand{\linopat}{m s O{} m d()}{%
  \IfBooleanTF{#2}{%
    \evallinop{\linop{#1}}*{#4}(#5)%
  }{%
    \evallinop{\linop{#1}}[#3]{#4}(#5)%
  }%
}

\NewDocumentCommand{\linfctlsat}{m s O{} m d()}{%
  \IfBooleanTF{#2}{%
    \evallinop{\linfctls{#1}}*{#4}(#5)%
  }{%
    \evallinop{\linfctls{#1}}[#3]{#4}(#5)%
  }%
}

\newcommand*{\diff}[2][1]{%
  \mathrm{d} #2\ifstrequal{#1}{1}{}{^{#1}}%
}

\newcommand*{\pdiff}[2][1]{%
  \partial #2\ifstrequal{#1}{1}{}{^{#1}}%
}

\NewDocumentCommand{\deriv}{s O{1} m m}{%
  \frac{%
    \mathrm{d}\ifstrequal{#2}{1}{}{^{#2}}\IfBooleanT{#1}{#4}
  }{%
    \diff[#2]{#3}
  }
  \IfBooleanF{#1}{#4}
}

\NewDocumentCommand{\derivat}{s O{1} m m m}{%
\left.
\IfBooleanTF{#1}{%
  \deriv*[#2]{#3}{#4}
}{%
  \deriv[#2]{#3}{#4}
}
\right|_{#3 = #5}
}

\NewDocumentCommand{\dderiv}{m m}{%
  \partial_{#1} #2
}

\NewDocumentCommand{\dderivat}{m m o m}{%
  \IfNoValueTF{#3}{%
    \dderiv{#1}{#2} \left( #4 \right)
  }{%
    \left.
    \dderiv{#1}{#2}
    \right_{#3 = #4}
  }
}

\NewDocumentCommand{\pderiv}{s O{1} m m}{%
  \frac{%
    \partial\ifstrequal{#2}{1}{}{^{#2}}\IfBooleanT{#1}{#4}
  }{%
    \pdiff[#2]{#3}
  }
  \IfBooleanF{#1}{#4}
}

\NewDocumentCommand{\pderivat}{s O{1} m m o m}{%
\left.
\IfBooleanTF{#1}{%
  \pderiv*[#2]{#3}{#4}
}{%
  \pderiv[#2]{#3}{#4}
}
\right|_{\IfNoValueTF{#5}{#3}{#5} = #6}
}

\NewDocumentCommand{\mpderiv}{s O{1} m m}{%
  \frac{%
    \partial\ifstrequal{#2}{1}{}{^{#2}}\IfBooleanT{#1}{#4}
  }{%
    #3
  }
  \IfBooleanF{#1}{#4}
}

\NewDocumentCommand{\mpderivat}{s O{1} m m m m}{%
\left.
\IfBooleanTF{#1}{%
  \mpderiv*[#2]{#3}{#4}
}{%
  \mpderiv[#2]{#3}{#4}
}
\right|_{#5 = #6}
}

\NewDocumentCommand{\mipderiv}{s m o m o}{%
  \IfNoValueTF{#3}{
    \mathrm{D}^{#2}
    \IfBooleanTF{#1}{\left[ #4 \right]}{#4}
    \IfNoValueF{#5}{\left( #5 \right)}
  }{
    \IfNoValueF{#5}{\left.}
    \frac{%
      \partial^{\lvert #2 \rvert}\IfBooleanT{#1}{#4}
    }{%
      \pdiff[#2]{#3}
    }
    \IfBooleanF{#1}{#4}
    \IfNoValueF{#5}{\right\vert_{#3 = #5}}
  }
}

\NewDocumentCommand{\jacobian}{o m o}{%
  \IfNoValueTF{#1}{%
    \mathrm{D} #2 \IfNoValueF{#3}{\left( #3 \right)}
  }{%
    \left.
      \mathrm{D} #2
    \right|_{#1\IfNoValueF{#3}{= #3}}
}
}

\NewDocumentCommand{\gradient}{o m o}{%
  \IfNoValueTF{#1}{%
    \nabla #2 \IfNoValueF{#3}{\left( #3 \right)}
  }{%
    \left.
      \nabla #2
    \right|_{#1\IfNoValueF{#3}{= #3}}
}
}

\NewDocumentCommand{\divergence}{m o}{%
  \operatorname{div} \left( #1 \right) \IfNoValueF{#2}{\left( #2 \right)}
}

\NewDocumentCommand{\hessian}{o m o}{%
  \IfNoValueTF{#1}{%
    H #2 \IfNoValueF{#3}{\left( #3 \right)}
  }{%
    \left.
      H #2
    \right|_{#1\IfNoValueF{#3}{= #3}}
}
}

\NewDocumentCommand{\laplaceop}{o m o}{%
  \IfNoValueTF{#1}{%
    \Delta #2 \IfNoValueF{#3}{\left( #3 \right)}
  }{%
    \left.
      \Delta #2
    \right|_{#1\IfNoValueF{#3}{= #3}}
}
}

\NewDocumentCommand{\rintegral}{o o m m}{%
  \int\IfNoValueF{#1}{_{#1}}\IfNoValueF{#2}{^{#2}} #4 \,\diff[1]{#3}%
}

\DeclareMathOperator*{\argmax}{arg\,max}

\makeatletter

\newcommand*{\@probsymbol}{\mathrm{P}}

\newcommand*{\@given}[1]{%
  \nonscript\:#1\vert
  \allowbreak
  \nonscript\:
  \mathopen{}}

\providecommand*{\given}{}

\DeclarePairedDelimiterXPP{\@prob}[1]{\@probsymbol}{(}{)}{}{%
  \renewcommand*{\given}{\@given{\delimsize}}%
  #1}

\newcommand*{\prob}[1]{\ifblank{#1}{\@probsymbol}{\@prob*{#1}}}

\makeatother

\makeatletter

\DeclarePairedDelimiterX{\@condrv}[1]{.}{.}{%
  \renewcommand*{\given}{\@given{\delimsize}}%
  #1}

\NewDocumentCommand{\condrv}{som}{%
  \IfBooleanTF{#1}{%
    \@condrv*{#3}
  }{%
    \IfNoValueTF{#2}{%
      \begingroup%
      \renewcommand*{\given}{\@given{}}%
      #3%
      \endgroup%
    }{%
      \@condrv[#2]{#3}%
    }
  }
}

\makeatother

\newcommand*{\rvec}[1]{{\bm{\mathrm{#1}}}}

\newcommand*{\rvq}{\rvec{q}}

\newcommand*{\rvu}{\rvec{u}}

\newcommand*{\rvy}{\rvec{y}}

\newcommand*{\rvepsilon}{\rvec{\epsilon}}

\NewDocumentCommand{\expectation}{o o m}{%
  \operatorname{\mathbb{E}}\IfNoValueF{#1}{_{#1\IfNoValueF{#2}{\sim #2}}} \left[ #3 \right]
}
\NewDocumentCommand{\covariance}{o o m m}{%
  \operatorname{Cov}\IfNoValueF{#1}{_{#1\IfNoValueF{#2}{\sim #2}}} \left[ #3, #4 \right]
}
\NewDocumentCommand{\variance}{o o m}{%
  \operatorname{\mathbb{V}}\IfNoValueF{#1}{_{#1\IfNoValueF{#2}{\sim #2}}} \left[ #3 \right]
}

\newcommand*{\gaussian}[2]{{\ensuremath{\operatorname{\mathcal{N}}\left(#1, #2\right)}}}

\NewDocumentCommand{\LkL}{m m o}{%
  #1 #2 \IfValueTF{#3}{#3}{#1}\dualop
}

\declaretheorem[style=plain]{theorem}

\declaretheorem[style=plain,sibling=theorem]{lemma}

\declaretheorem[style=definition,sibling=theorem]{definition}

%% file: preamble/07_draft.tex
\usepackage[
  obeyDraft,
  textsize=tiny,
  figwidth=0.99\linewidth,
]{todonotes}

%% file: preamble/08_paper_macros.tex
\newcommand{\gmp}{\rvu}

\newcommand{\gmpdim}{d}
\newcommand{\gmpA}{\mA}
\newcommand{\gmpb}{\vb}
\newcommand{\gmpnoise}{\rvq}
\newcommand{\gmpnoisecov}{\mQ}
\newcommand{\gmpmean}{\vmu}
\newcommand{\gmpcov}{\mSigma}

\newcommand{\obs}{\vy}
\newcommand{\obsrv}{\rvy}
\newcommand{\obsdim}{n}
\newcommand{\obsH}{\mH}
\newcommand{\obsnoise}{\rvepsilon}
\newcommand{\obsnoisecov}{\mLambda}

\newcommand{\cakfstate}{\hat{\gmp}^f} % updated state RV
\newcommand{\cakfpstate}{\hat{\gmp}^{f-}}  % predictive state RV
\newcommand{\cakfpmean}{\hat{\vm}^-}  % predictive mean
\newcommand{\cakfpdd}{\hat{\mM}^-}  % predictive covariance downdate
\newcommand{\cakfpcov}{\hat{\mP}^-}  % predictive covariance
\newcommand{\cakfmean}{\hat{\vm}}  % updated mean
\newcommand{\cakfdd}{\hat{\mM}}  % updated downdate (mind = blown)
\newcommand{\cakfcov}{\hat{\mP}}  % updated covariance
\newcommand{\cakftdd}{\hat{\mM}^+}  % truncated covariance downdate
\newcommand{\cakfw}{\hat{\vw}}

\newcommand{\cakfW}{\hat{\mW}}
\newcommand{\cakfacts}{\mS}  % Matrix of all PLS actions per time step
\newcommand{\cakfgram}{\hat{\mG}}

\newcommand{\cakfprojobs}{\check{\obs}}
\newcommand{\cakfprojobsrv}{\check{\obsrv}}
\newcommand{\cakfprojobsdim}{\check{n}}
\newcommand{\cakfprojH}{\check{\obsH}}
\newcommand{\cakfprojobsnoise}{\check{\obsnoise}}
\newcommand{\cakfprojobsnoisecov}{\check{\obsnoisecov}}
\newcommand{\cakfprojgram}{\check{\mG}}
\newcommand{\cakfprojgramlsqrt}{\check{\mV}}

\newcommand{\kfpstate}{\gmp^{f-}}  % predictive state
\newcommand{\kfpmean}{\vm^-}  % predictive mean
\newcommand{\kfpcov}{\mP^-}  % predictive covariance
\newcommand{\kfstate}{\gmp^f}  % Updated state RV
\newcommand{\kfmean}{\vm}  % updated mean
\newcommand{\kfcov}{\mP}  % updated covariance

\newcommand{\kfgram}{\mG}

\newcommand{\kfgain}{\mK}

\newcommand{\ntraindata}{n}

\newcommand{\ntimesteps}{n_t}
\newcommand{\nspatialdata}{n_x}

\newcommand{\idxdtime}{k}
\newcommand{\idxdtimefinal}{K}

\newcommand{\E}{\mathbb{E}}

%% file: preamble/99_end.tex
\usepackage[
  capitalize,
  nameinlink,
  noabbrev,
]{cleveref}

\makeatletter
\if@cref@capitalise
  \crefname{assumption}{Assumption}{Assumptions}
\else
  \crefname{assumption}{assumption}{assumptions}
\fi
\makeatother

%% file: 00_abstract.tex
Due to their explicit priors and ability to model uncertainty, Bayesian methods have played a major role in dynamical latent variable modeling of single-cell neural recordings. 
However, modern-sized datasets have made overparameterized deep networks the preferred methods of choice due to their predictive power and favorable computational scaling. While many posterior approximations exist, all incur approximation errors. 
Recent work accounts for this error in the form of computational uncertainty but comes at the cost of quadratic complexity and assumes fixed model hyperparameters. 
Here we extend this development to model selection, including a novel training loss and optimization scheme, which yields tractable inference in large state-spaces. 
We introduce a framework, the Computation-Aware State-Space Model (CASSM), specifically designed for the \emph{scale-imbalanced regime}, where the number of trials is significantly lower than the number of recorded neurons. 
In this regime, for both synthetic and real data, we show that our method is competitive with data-hungry deep networks, with significantly improved uncertainty calibration over previous attempts to scale Bayesian methods. 
Our experiments provide a roadmap to neuroscience researchers in choosing from a host of potential dynamical latent variable models given key dataset properties and constraints. 

%% file: 01_introduction.tex
\section{Introduction}
\label{sec:introduction}
%1) X (+define X if not obvious) is an important problem 
 While behaviors are governed by large populations of neurons, many such processes may be well-described by low-dimensional manifold structure \citep{Gallego2017neural}. Neural interactions can be summarized as a set of latent variables evolving in a low-dimensional space, allowing neuroscientists to decode internal representations and neural dynamics with computational models. However, with the largest datasets now far beyond the typical tens or hundreds neurons at varying temporal resolutions \citep{Siegle2021Spiking,Steinmetz2019Distributed}, dynamical latent variable models are becoming computationally strained. 
 
 As our understanding of neural dynamics shifts with this increase in data, the manifold hypothesis may reveal itself as more of a computational convenience than scientific reality. ``Low-dimensional systems'' might be so only relative to the scale of systems with billions of components. \citep{Gauthaman2025scaleFree}, for example, posits that dimensionality reduction may be the wrong approach to modeling neural representations, showing that across the human visual cortex, dimensionality is unbounded and scales with dataset size. 

%2) The core challenges are this and that. 
To predict neural dynamics from spiking data is a precarious balance of model capacity and interpretability. As we seek to understand the structure of neural representations, an ideal model would both predict new data well \emph{and} directly map its underlying components to experimental objects of interest, like a proposed differential equation. A Bayesian method typical of dynamical latent variable models like Gaussian Process Factor Analysis (GPFA) \citep{Yu2008GPFA} can flexibly incorporate prior knowledge and quantify predictive uncertainty, but the model's cubic complexity with respect to temporal resolution is a limiting factor for modern datasets. 

\begin{figure}[t]
  \centering
  \includegraphics[width=\columnwidth]{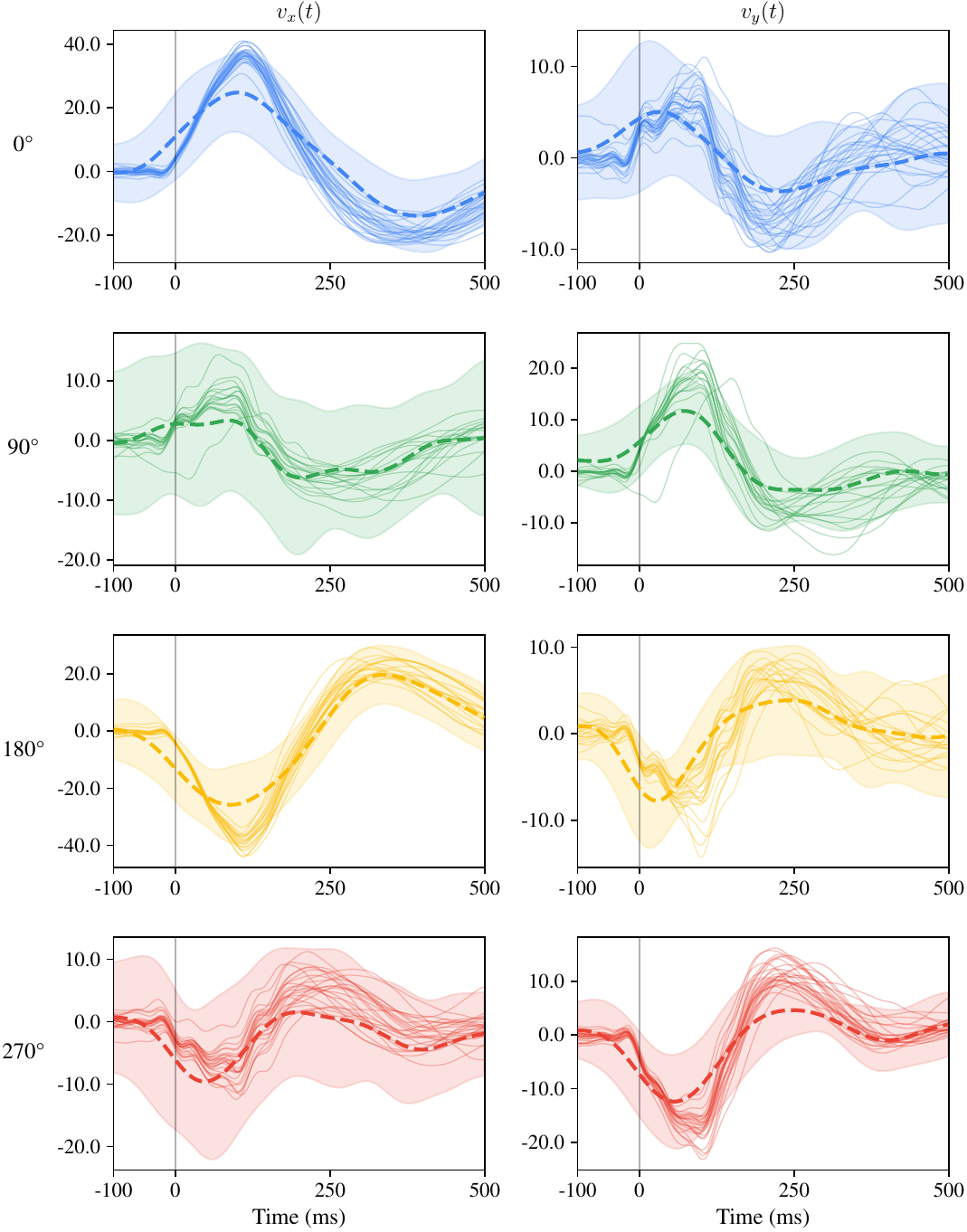}
  \caption{Hand velocities in the $x-y$ plane for four different directions ($0^\circ$, $90^\circ$, $180^\circ$, $270^\circ$) for the Area 2 Bump dataset, where a primate is bumped from different angles during a reaching task. The mean predictions and standard deviations from CASSM are shown in the dashed line and filled areas, respectively. Kinematics are obtained by sampling from the CASSM filtering distribution after training and averaging over the linear decoder fits.}
  \label{fig:velocity}
\end{figure}

In contrast, nonlinear overparameterized deep algorithms like Latent Factor Analysis via Dynamical Systems (LFADS) \citep{Pandarinath2018LFADS} scale more favorably and have seen success across many applications, from calcium imaging modalities \citep{Zhu2021Deepinference} to intracortical brain-computer interfaces \citep{Karpowicz2022BCI}, yet such models may have limited ability to quantify uncertainty and, most importantly, require large numbers of trials to offset their relative lack of inductive bias. 

We term this data-limited setting the \emph{scale-imbalanced regime}, where the number of trials is far less than the number of neurons. We consider this regime to be the future of neuroscience. In the context of the human brain, for example, hundreds of billions of neurons cannot reasonably yield hundreds of billions of trials without heavy data augmentation. In other words, compute-optimal scaling laws for neuroscience models will necessarily lag far behind other modalities like language \citep{hoffmann2022training}. 

%3) Previous work on X has addressed these with Y, but the problems with this are Z. 

\subsection{Previous Work and Limitations}

Previous work has addressed the scaling issues of Bayesian methods with approximation methods. Typically, this approximation exploits the low-dimensional structure of neural dynamics. Sparse variational inference and inducing point methods \citep{duncker2018temporal,Jensen2021Scalable} are two such examples. In particular, bGPFA \citep{Jensen2021Scalable} develops a scalable Bayesian version of GPFA.  However, while one of the main advantages of a Bayesian framework is uncertainty quantification, this approximation then leaks into uncertainty (and mean) estimates without scrutiny. For example, one typical manifestation of this leakage in Bayesian optimization is overconfidence or ``variance starvation'' \citep{Wang2018batched,Wenger2022PosteriorComputational}. 

By abstracting this approximation away from inference, two neuroscientists could have the same model, prior knowledge, data, even hardware and computing resources, but reach different conclusions without accounting for the biases introduced by a seemingly-heuristic approximation. The Computation-Aware Kalman Filter (CAKF) \citep{Pfoertner2024ComputationAware} has addressed the quantification of approximation error with the notion of \emph{computational uncertainty}, but not model selection. Prespecification of the dynamics and observation models is a major limitation in neuroscientific applications where these are \emph{a priori} unknown and must be learned from data.

\begin{figure}
  \centering
  \includegraphics[width=0.4\textwidth]{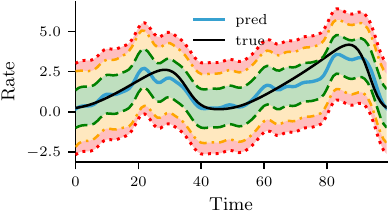}
  \caption{As the amount of computation decreases and posterior approximation increases, the uncertainty bands around the neural trajectory estimates of CASSM widen: (a) in green, we see that the Kalman smoother uncertainty estimates in the case of no approximation of a $30$-dimensional state-space; (b) in orange, we see the added uncertainty from projecting this $30$-dimensional system to a $10$-dimensional space; (c) in red, we see the further added uncertainty from projecting to a $2$-dimensional space.}
  \label{fig:bands}
 \end{figure}

%4) In this work we do W (?). 

\subsection{Our Approach}

In this work, we introduce a probabilistic numerical method, the Computation Aware State-Space Model (CASSM), that learns a low-dimensional projection of Kalman-filtered dynamics by matching the projected and computationally infeasible true posterior \citep{Wenger2024computationawaregp}. In particular, we optimize a decomposition of the ELBO into a data-fit term and a divergence term that penalizes deviation from the exact filtering distribution. Extending core ideas of the CAKF \citep{Pfoertner2024ComputationAware} to model selection, this projection yields precise estimates of the uncertainty under incomplete computation, and increases uncertainty relative to the amount of approximation. We show this phenomenon in Figure \ref{fig:bands}, a decomposition that is not possible for previous approximation methods like \citep{duncker2018temporal,Jensen2021Scalable}. We show that our method can be mathematically interpreted as a principal components analogue wrapped in a state-space model, which benefits from but does not explicitly require low-dimensional manifold structure as the number of simultaneously recorded neurons increases. Instead, we assume that an idealized posterior exists, but limited computational capacity prohibits direct access to this posterior. Our method is \emph{not} intended for generalized use cases: it is highly specialized for the scale-imbalanced regime, but in this regime, it is highly competitive. 

By translating the Gaussian process representation to a state-space representation, we also introduce a novel, natural technique to explicitly encode a \emph{spatial prior} over neurons, allowing the user to specify spatial coordinates or otherwise differentiate neurons by cell type and location, a feature which is only implicit in GPFA, LFADS, or their variants.
%5) This has the following appealing properties and our experiments show this and that. 

 \textbf{Contribution}
We introduce the Computation-Aware State-Space Model (CASSM), a probabilistic numerical framework that: 
\begin{enumerate}[itemsep=-0.5ex,topsep=0ex,partopsep=0ex,label=(\arabic*)]
  \item \emph{enables model selection} with a novel loss function that enables efficient hyperparameter optimization and preserves the mathematical interpretation of computational uncertainty for inference in the CAKF \citep{Pfoertner2024ComputationAware}
\end{enumerate}

%% file: 02_background.tex
\section{Background}

% Introduce SSMs

\subsection{State-Space Models}
\label{sec: ssms}

A state-space model (SSM) is one particular choice of representation for dynamical systems that has gained significant popularity in recent years for sequence modeling in natural language processing \citep{Gu2023MambaLinearTime}. In this framework, we wish to infer latent state variables $\gmp_\idxdtime$ from a given set of noisy observations $\set{\obs_\idxdtime}_{\idxdtime = 1}^{\ntimesteps}$ through the following two-stage transformation:
\begin{subequations}\label{eq:state-space-model}
  \begin{align}
  \label{eq:state-transition}
  \gmp_\idxdtime
  &=
  \gmpA_{\idxdtime-1}\gmp_{\idxdtime-1}
  +\gmpb_{\idxdtime-1}
  +\gmpnoise_{\idxdtime-1}
  \in \R^\gmpdim,
  \\
  \label{eq:observation-model}
  \obsrv_\idxdtime
  &=
  \obsH_{\idxdtime}\gmp_\idxdtime
  +\obsnoise_\idxdtime
  \in \R^{\obsdim}.
  \end{align}
  \end{subequations}
which we refer to as the \emph{dynamics equation} and \emph{observation equation}, respectively. Unlike the setting of the CAKF \citep{Pfoertner2024ComputationAware}, we consider $\Theta_\idxdtime = \{ \gmpA_\idxdtime, \gmpb_\idxdtime, \gmpnoise_\idxdtime, \obsH_\idxdtime, \obsnoise_\idxdtime \}_{\idxdtime = 1}^{\ntimesteps}$ unknown in addition to $\set{\gmp_\idxdtime}_{\idxdtime = 1}^{\ntimesteps}$, though we do also consider the setting of linear Gaussian state-space models. Under the conditions of a Gauss-Markov stochastic process, exact Bayesian inference of the conditional distributions $\gmp_\idxdtime \mid \obsrv_{1: \idxdtime}$ can be done using the Kalman filter. 

We include a more in-depth discussion of filtering in Appendix \ref{filtering}, but at a high level, the exact Kalman filter proceeds as follows. First, we assume initial prior (unconditional) state $\gmp_0 \sim \gaussian{\gmpmean_0}{\gmpcov_0}$. Next, we assume the \emph{process noise model} $\gmpnoise_{\idxdtime-1} \sim \gaussian{\vec{0}}{\gmpnoisecov_{\idxdtime-1}}$ and \emph{measurement noise model} $\obsnoise_\idxdtime \sim \gaussian{\vec{0}}{\obsnoisecov_\idxdtime}$. 
The algorithm recursively computes the filtering distribution \(
\kfstate_\idxdtime
\defeq \ps{\gmp_\idxdtime \mid \obsrv_{1:\idxdtime} = \obs_{1:\idxdtime}}
\sim \gaussian{\kfmean_\idxdtime}{\kfcov_\idxdtime}
\)
 by first computing the \emph{pre-update predictive distribution} $\kfpstate_\idxdtime$:
\begin{align*}
  \kfpmean_\idxdtime & \defeq \gmpA_{\idxdtime-1} \kfmean_{\idxdtime-1}                                                   \\
  \kfpcov_\idxdtime  & \defeq \gmpA_{\idxdtime-1} \kfcov_{\idxdtime-1} \gmpA_{\idxdtime-1}\T + \gmpnoisecov_{\idxdtime-1}
\end{align*}
for
\(
\kfpstate_\idxdtime
\defeq \ps{\gmp_\idxdtime \mid \obsrv_{1:\idxdtime-1}}
\sim \gaussian{\kfpmean_\idxdtime}{\kfpcov_\idxdtime}
\)
in the \emph{predict step} and the moments

\begin{align}
  \kfmean_\idxdtime & \defeq \kfpmean_\idxdtime + \kfpcov_\idxdtime \obsH_\idxdtime \kfgram_\idxdtime^{-1} (\obsrv_\idxdtime - \obsH_\idxdtime \kfpmean_\idxdtime) \label{eq:kalman_filtering_mean} \\
  \kfcov_\idxdtime  & \defeq \kfpcov_\idxdtime - \kfpcov_\idxdtime \obsH_\idxdtime \kfgram_\idxdtime^{-1} \obsH_\idxdtime\T \kfpcov_\idxdtime \label{eq:kalman_filtering_cov}
\end{align}

of
\(
\kfstate_\idxdtime
\defeq \ps{\gmp_\idxdtime \mid \obsrv_{1:\idxdtime}}
\sim \gaussian{\kfmean_\idxdtime}{\kfcov_\idxdtime}
\)
in the \emph{update step}, where the innovation matrix $\kfgram_\idxdtime \defeq \obsH_\idxdtime \kfpcov_\idxdtime \obsH_\idxdtime\T + \obsnoisecov_\idxdtime$.

If the initial state, process noise, and measurement noise are pairwise independent, then $\gmp_\idxdtime \sim \gaussian{\gmpmean_\idxdtime}{\gmpcov_\idxdtime}$, where: 
\begin{subequations}
  \begin{align}
  \label{eq:posterior-mean}
  \gmpmean_\idxdtime &\defeq \gmpA_{\idxdtime - 1} \gmpmean_{\idxdtime - 1} + \gmpb_{\idxdtime - 1} \\
  \label{eq:posterior-cov}
  \gmpcov_\idxdtime &\defeq \gmpA_{\idxdtime - 1} \gmpcov_{\idxdtime - 1} \gmpA_{\idxdtime - 1}\T + \gmpnoisecov_{\idxdtime - 1}
  \end{align}
  \end{subequations}

in the \emph{prior update step}. 

\subsection{Tradeoffs in LGSSMs}
\label{sec: ssms-advantage}
Standard regression approaches like Gaussian Process Factor Analysis have cubic cost \(\mathcal{O}{(\ntimesteps^3)}\) for temporal resolution $\ntimesteps$, which requires computational improvisations on modern datasets. 
 Instead of artificially chunking time series into pseudo-trials and treating these as independent samples, one can leverage the inherent temporal structure of neural dynamics by performing Bayesian filtering and smoothing (BFS), which has linear time complexity \(\mathcal{O}(\ntimesteps)\) \citep{Sarkka2023BayesianFiltering}.

 However, there are multiple problems with BFS, as we: (P1) maintain the $\mathcal{O} (\nspatialdata^3)$ time and $\mathcal{O} (\nspatialdata^2)$ memory of the classic regressions for spatial resolution $\nspatialdata$ due to the inversion and storage of the innovation matrix $\kfgram_\idxdtime$; (P2) require $\mathcal{O} (\gmpdim^2)$ memory for covariances $\kfcov_\idxdtime$ and $\kfpcov_\idxdtime$; (P3) $\Theta_k$ must be pre-specified, and learning these quantities is infeasible when $n$ and $d$ are large without significant structual modifications. The CAKF addresses (P1) and (P2), while CASSM addresses (P3).

\subsection{Computation-Aware Kalman Filtering}

Core ideas of efficient filtering have been introduced in \citep{Pfoertner2024ComputationAware} and \citep{Pnevmatikakis2014FastKalman}. Central to both works are the assumed low-rank structure on the posterior covariance matrix. \citep{Pnevmatikakis2014FastKalman} approximates the posterior covariance using the Woodbury identity, while the CAKF \citep{Pfoertner2024ComputationAware} introduces a fixed projection of the data.  In either case, the posterior covariance can be approximated quite accurately by a low-rank perturbation of the prior equilibrium (stationary) covariance of the state $\kfstate_\idxdtime$.  To reduce time and memory complexity of the update step in CASSM, we project the observation as in the CAKF:
\(
\cakfprojobsrv_\idxdtime
\defeq \cakfacts_\idxdtime\T \obsrv_\idxdtime
\)
but with a \emph{learned} policy $\cakfacts_\idxdtime \in \R^{\ntraindata \times \cakfprojobsdim}$ and $\cakfprojobsdim \ll \ntraindata$. The updated observation equation is:
\begin{equation}
  \label{eqn:proj-observation}
  \cakfprojobsrv_\idxdtime = \underbracket[0.1ex]{\cakfacts_\idxdtime\T \obsH_\idxdtime}_{\rdefeq \cakfprojH_\idxdtime} \gmp_\idxdtime + \underbracket[0.1ex]{\cakfacts_\idxdtime\T \obsnoise_\idxdtime}_{\rdefeq \cakfprojobsnoise_\idxdtime} \in \R^{\cakfprojobsdim}
\end{equation}
Storage of $\cakfgram_\idxdtime$ is reduced to $\mathcal{O} (\cakfprojobsdim^2)$ and its inversion is reduced to $\mathcal{O} (\cakfprojobsdim^3)$. In other words, the algorithm only scales prohibitively with respect to the intrinsic dimensionality of the data, solving (P1). The resulting predictive and filtering moments $\set{\cakfpmean_\idxdtime, \cakfpcov_\idxdtime}_{\idxdtime = 1}^{\ntimesteps}$ and $\set{\cakfmean_\idxdtime, \cakfcov_\idxdtime}_{\idxdtime = 1}^{\ntimesteps}$ are approximations of the filtering distributions. The updated filtering covariance yields:
\begin{align*}
  \cakfcov_\idxdtime = \cakfpcov_\idxdtime - \cakfpcov_\idxdtime \cakfprojH_\idxdtime\T \cakfprojgramlsqrt_\idxdtime (\cakfpcov_\idxdtime \cakfprojH_\idxdtime\T \cakfprojgramlsqrt_\idxdtime)\T
\end{align*}
for $\cakfprojgramlsqrt_\idxdtime \cakfprojgramlsqrt_\idxdtime\T = \cakfprojgram_\idxdtime\inv$. We can see that the projection alone does not solve (P2), as $\cakfcov_\idxdtime$ remains an $\mathcal{O} (\gmpdim^2)$ object. Instead of considering the update to $\cakfcov_\idxdtime$, we can consider low-rank perturbations around the prior process noise. The update-predict recursions have an equivalent low-rank downdate structure:
\begin{equation}\label{eq:cov-update}
  \begin{aligned}
    \cakfpcov_{\idxdtime} &= \gmpcov_{\idxdtime} - \cakfpdd_{\idxdtime} (\cakfpdd_{\idxdtime})^\top,\\
    \cakfcov_{\idxdtime}  &= \gmpcov_{\idxdtime} - \cakfdd_{\idxdtime}\,\cakfdd_{\idxdtime}^\top.
  \end{aligned}
  \end{equation}
where \(\cakfpdd_\idxdtime = \gmpA_{\idxdtime - 1} \cakfdd_{\idxdtime - 1}\) and \(\cakfdd_\idxdtime  = (\cakfpdd_\idxdtime \ \ \ \cakfpcov_\idxdtime \cakfprojH_\idxdtime\T \cakfprojgramlsqrt_\idxdtime)\). Thus, in our update step, we only store the ``skinny'' low-rank square roots $\cakfpdd_\idxdtime, \cakfdd_\idxdtime \in \R^{{d \times r}}$. The latent covariances can be lazily accessed with matrix-vector products $\gmpA_{\idxdtime -1} \vw$, $\gmpcov_\idxdtime \vw$, and $\obsH_\idxdtime\vw$. By imposing kernel structure on the prior covariance $\gmpcov\left[ (t_1, \mathbf{x}_1), (t_2, \mathbf{x}_2)\right]$, we decrease the number of trainable parameters and perform these multiplications without explicitly forming $\cakfpcov_\idxdtime, \cakfcov_\idxdtime$ in memory. This solves (P2), reducing memory requirements from $\mathcal{O} (\gmpdim^2)$ to $\mathcal{O} (\gmpdim r )$. We include more filter-specific details for the CAKF in Appendix \ref{filtering}. 

However, this filtering structure does not address selection of hyperparameters $\Theta_\idxdtime$ (P3). Furthermore, choice of action can influence runtime and memory considerations. Previous work either suggests but does not implement strategies for the model selection task or assumes that hyperparameters are already known. To address principled and fast hyperparameter selection, we first introduce a new interpretation of the policy.

 % Prior information / decision making -> uncertainty quantification via probabilistic models?

% Kalman Filters are efficient solvers under Gauss-Markov, XYZ assumptions

% DKFs and LFADS, compare/contrast with KFs

% Challenges (matrix inversion)

% Start to head into loss_deprecated function in Model section

%% file: 03_model.tex
\section{CASSM}
\label{sec:model}

% loss function

% Discuss the architecture + training process

% put Kalman algorithms in the appendix and highlight new additional objects (low-dimensional projection, etc.)

% matrix-free notes here

\subsection{Actions as Principal Components Analysis}
\label{sec: pca}

Model selection enables a data-driven interpretation of the learned policy $\cakfacts_{\idxdtime}$. In particular, we establish a connection to Principal Components Analysis (PCA) via Theorem \ref{entropy} from an information-theoretic perspective. We include the proof in Appendix \ref{sec: pca}.

\begin{theorem}[Greedy Posterior Entropy Minimization]
  \label{entropy}
  Given a sequence of actions $\cakfacts_1, \ldots \cakfacts_{\idxdtime-1}$, the action $\cakfacts_{\idxdtime}$ that maximally reduces the latent uncertainty given by the entropy of the posterior update is given by the top $\cakfprojobsdim$ eigenvectors of the innovation covariance $\kfgram_\idxdtime$ 
\end{theorem}

% look at matern.jl file 
 Geometrically, the optimal policies are the directions of greatest expected uncertainty in the measurement residual $\vr_\idxdtime := \obsrv_\idxdtime - \obsH_\idxdtime \kfpmean_\idxdtime$ during the update step (Eq. \ref{eq:kalman_filtering_mean}). If GPFA \citep{Yu2008GPFA} is a Factor Analysis model wrapped in a Gaussian Process (GP) regression, then CASSM is essentially a PCA analogue wrapped in a state-space model. 
 
 \subsubsection{Subspace matching} 
\begin{figure}
  \centering
  \includegraphics[scale=0.65]{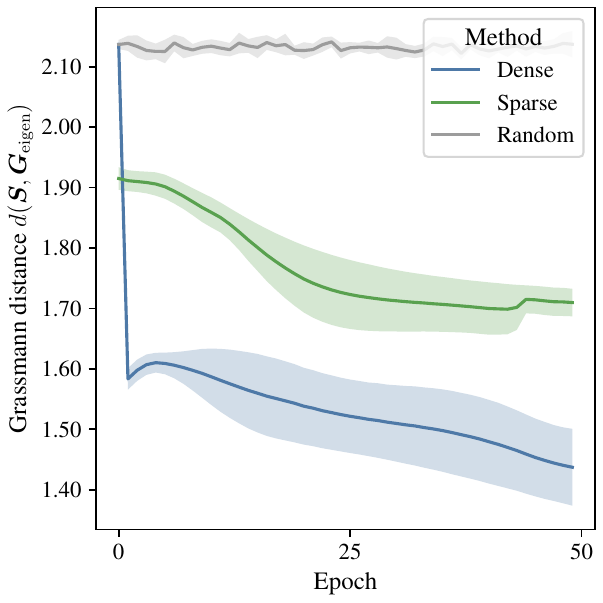}
  \caption{Grassmann distance between the learned $\cakfacts_\idxdtime$ (dense and sparse variations) and the top eigenvectors of $\kfgram_\idxdtime$. The distance decreases during training, indicating that the learned policy converges to the optimal theoretical policy at small neuron scales where PCA is possible. The random policy is a random projection that is kept fixed throughout training. }
  \label{fig:grassmann_distance}
 \end{figure}

We verify this theorem empirically in Figure \ref{fig:grassmann_distance}. To validate our theorem, we fix the model hyperparameters $\Theta_k$ obtained from a classical Kalman Filter that was trained with a log-marginal likelihood loss \citep{Sarkka2023BayesianFiltering}, and then train $\cakfacts_\idxdtime$ for dense and sparse actions. While both flavors incur linear memory requirements, dense actions naively incur quadratic time complexity in the state-space dimensionality $n$. To decrease computational cost and fully leverage hardware acceleration, we impose a sparse block structure on $\cakfacts_\idxdtime$. For each $j \in [\cakfprojobsdim]$, a block is a column vector $\vs_j \in \R^{k \times 1}$ with $k = n / \cakfprojobsdim$ entries such that the total number of trainable action parameters equals the number of neurons. This sparsity constraint reduces the time complexity of posterior inference and model selection to linear in the state-space dimensionality.  
 
 In Figure \ref{fig:grassmann_distance}, we compute the top five eigenvectors for a small $n = 50$ system of neurons where PCA of $\kfgram_\idxdtime$ is trivial. We then compute the Grassmann distance between these eigenvectors and either of the sparse or dense actions. Compared to a choice of randomly fixed actions, optimizing these actions with our proposed hyperparameter optimization in Section \ref{sec: loss-function} produces an alignment with optimal action choice minimizing posterior entropy, where the dense actions align more closely than the sparse actions, which are more heavily constrained in their class of directions.

 \subsubsection{Utility of the Theorem}
 
 While Theorem \ref{entropy} gives some intuition for the function of $\cakfacts_\idxdtime$ and our overall framework, its practicality is limited as it shares the $\mathcal{O} (\nspatialdata^3)$ time and $\mathcal{O} (\nspatialdata^2)$ memory for computing the original GP posterior. In the regime where this is possible, approximation is not necessary. Even if there existed structural exploits on $\kfgram_\idxdtime$ in the infeasible regime, the assumption of fixed model hyperparameters $\Theta_k$ is unwieldy, regardless. Because $\gmpA_{\idxdtime}$ is assumed to be stable in CASSM, which is a common assumption \citep{Paninski2010FastKalmanDendritic} \citep{Bonnabel2012GeometryLowRank} in general, long-ago observations are forgotten for $k \to \infty$ as the double multiplication of $\gmpA_{\idxdtime}$ in Eq. \eqref{eq:cov-update} causes exponential decay in the downdate. 
 
 %: as $\idxdtime \to \infty$, the covariance $\cakfpcov_\idxdtime$ is expected to decrease in rank by nature of the system, so
 
 Thus, an EM-style strategy as hinted at in \citep{Pnevmatikakis2014FastKalman} wastes gradient computation of $\cakfacts_\idxdtime$ with respect to noisier estimates of $\Theta_k$ in earlier timesteps. Ideally, we would optimize the actions alongside the hyperparameters end-to-end, such that the training loss for model selection defines what data projections are informative. 

\subsection{Model Selection in CASSM}
\label{sec: loss-function}

A core contribution of our framework is an efficient loss function for model selection. \citep{Sarkka2023BayesianFiltering} proposed the negative log-likelihood (NLL) for this task in the exact filter. Given the projected models, a reasonable solution is to simply evaluate the NLL in the approximate space. Empirically, the performance of this solution degrades rapidly with respect to the projection compression, irrespective of the data's intrinsic dimensionality. Of course, this solution only considers the update $\hat{\vr}_\idxdtime := (\cakfprojobs_\idxdtime - \cakfprojH_\idxdtime \kfpmean_\idxdtime) $. If the prediction $\obsH_\idxdtime \kfpmean_\idxdtime$ were computationally infeasible, however, then summarizing the activity of large populations of neurons would be a futile task in general. 

Another proposed solution is the application of the low-rank determinant lemma to compute the marginal log-likelihood in approximate space that is mentioned, but not implemented, in \citep{Pnevmatikakis2014FastKalman}. However, this is quite similar to the aforementioned NLL in the approximate space, as we are maximizing likelihood under the wrong posterior. Instead, we wish to maximize likelihood under true model and penalize deviation from the true posterior as in \citep{Wenger2024computationawaregp}.
Specifically, we utilize a regularization factor to minimize the divergence between $\kfstate_\idxdtime$ and the distribution of the approximation space $\cakfstate_\idxdtime$. While the loss ultimately balances multiple objectives, including the minimization of the projected residual, most crucially the CASSM loss predicts neural activity in the original space using the Mahalanobis distance $d_M (\obsrv_\idxdtime , \obsH_\idxdtime \kfpmean_\idxdtime, \obsnoisecov_\idxdtime)$. We formalize this idea in Definition \ref{def:loss-fnc}.

\begin{definition}[CASSM Objective]
  \label{def:loss-fnc}
  \begin{equation*}
    \begin{aligned}
      \mathcal{L} (\Theta)
      &= \sum_{\idxdtime \in [\idxdtimefinal]}
      \underbracket[0.1ex]{\E_{q}\left[\log p (\obsrv_\idxdtime \mid \gmp_\idxdtime ) \right]}_{\text{data-fit term}}
      \\
      &\quad + \sum_{\idxdtime \in [\idxdtimefinal]}
      \underbracket[0.1ex]{D_{KL} \left( q(\gmp_\idxdtime \mid \obsrv_{1:\idxdtime}) \parallel p(\gmp_\idxdtime \mid \obsrv_{\idxdtime - 1}) \right)}_{\text{divergence from exact filtering distribution}}.
    \end{aligned}
  \end{equation*}
\end{definition}

The objective is an evidence lower bound (ELBO) on the marginal likelihood $\log p(\obsrv_{1:\idxdtimefinal} \mid \Theta)$, where the approximate distribution $q(\gmp_\idxdtime \mid \obsrv_{1:\idxdtime})$ plays the role of a variational posterior. When $\obsdim = \cakfprojobsdim $, the divergence term vanishes and CASSM reduces to a low-rank Kalman Filter with model selection, similar to \citep{Paninski2010FastKalmanDendritic} \citep{Pnevmatikakis2014FastKalman}. Although both the expectation and divergence terms of the Evidence Lower Bound (ELBO) involve computationally intractable terms, these problematic terms cancel out when combined. We include a derivation of the specific numerical form and a discussion of other practical considerations in Lemma \ref{formnumeric} in Appendix \ref{sec:loss-function-derivation}.

\subsection{Training Procedure and Numerical Considerations}
\label{sec: procedure}

\subsubsection{Benefits of Kernel Structure}

The prior process update (Eqs. \ref{eq:posterior-mean}, \ref{eq:posterior-cov}), in the case of a linear time-invariant (LTI) system, induces a recursion that involves computing a series of matrix powers, which is computationally wasteful and sometimes numerically unstable. In deterministic settings, alternative SSM frameworks like H3 \citep{Fu2022hungry} overcome this difficulty by implementing a global convolutional form. Similarly, in the case of an LTI stochastic differential equation (SDE), the CASSM prior update can also be computed in closed form, avoiding this brute force recursion. Other low-rank LTI filter implementations that impose various structures on $\gmpA_\idxdtime$ like block-tridiagonal or symmetric sparse matrix structure cannot leverage this form and avoid the recursion, which is one of the benefits of the kernel structure. 

We show how this efficient reformulation has been done \citep{Sarkka2010gp} without approximation for certain kernel classes in Appendix \ref{sec:lti-sde}. In particular, the reformulation makes our connection with Gaussian Processes explicit, and exemplifies why GPFA is such a foundational work in models of latent neural trajectories. 
% however, this is a potential area of future research that lfads and gpfa don't have

\subsection{Spatial Priors}

While Bayesian filters and smoothers are efficient solvers of spatiotemporal regression problems, they also allow for more explicit representations of prior scientific knowledge. In GPFA and its modern variants \citep{Jensen2021Scalable}, spatial interaction is only modeled through a single loading matrix $\mC_\idxdtime$. In LFADS, spatial structure is learned via the generator and readout functions. Spatial interaction in CASSM is not only modeled through the learned measurement function $\obsH_\idxdtime$, but in the prior process covariance decomposition of $ \gmpcov\left[ (t_1, \mathbf{x}_1), (t_2, \mathbf{x}_2)\right] = \gmpcov^{(\mathbf{x})}(\mathbf{x}_1, \mathbf{x}_2) \cdot \gmpcov^{(t)} (t_1, t_2) $ through $\gmpcov^{(\mathbf{x})}(\mathbf{x}_1, \mathbf{x}_2)$. While the chosen kernel class determines $\gmpcov^{(t)} (t_1, t_2)$, which is fixed during training and inference, $\gmpcov^{(\mathbf{x})}(\mathbf{x}_1, \mathbf{x}_2)$ is user-determined.

This \emph{spatial prior} allows the user to explicitly specify neuron locations or cell types before training, if desired. If this input is unavailable, as is the case for certain datasets in our experiments like the synthetic Lorenz dataset, one can simply learn a position vector from data. We simply learn 3D coordinates for each neuron in all experiments, but if even this is too much of a refinement, one could simply pass neuron labels or brain regions into this vector instead. 

% spatial prior here

\subsubsection{Preprocessing}

The Kalman Filter and its low-rank variants specify Gaussian likelihoods, which we found is best served by two specific preprocessing steps. First, the discrete, binary-valued input spike trains (see Figure \ref{fig:lorenz-generative}) are convolved with a Gaussian kernel to obtain real-valued smoothed firing rates. While one of the original benefits of GPFA is to unify the spike-smoothing and dimensionality-reduction operations in a common probabilistic framework, we can simply add the kernel standard deviation to the computational graph and learn this parameter from data.  After convolution, we then apply the Anscombe variance stabilizing transform:
\begin{align*}
  A(x) := 2 \sqrt{x + \frac{3}{8}}
\end{align*}

to convert (pseudo) Poissonian data to approximately Gaussian data and ensure firing rate comparability across trials.  

\subsubsection{Model Selection and Numerical Stability}

Many low-rank filters propose an optimal truncation step for the low-rank downdate $\cakfdd_{\idxdtime}$ to avoid accumulation of $\gmpdim \times r $ matrices over time, which can easily exceed the state-space dimensionality if $T$ is large. This is typically accomplished with a singular value decomposition on $\cakfdd_{\idxdtime}$ and dropping the subspace corresponding to the smallest singular values $s_i$. However, when introducing hyperparameter optimization and backpropogating through time, the truncation step becomes numerically unstable due to the gradient computation in the backward pass, which depends the matrix \citep{Wang2021RobustDifferentiableSVD}:
\begin{align*}
  \mF_{ij} =
\begin{cases}
\dfrac{1}{s_j^{2} - s_i^{2}}, & i \neq j, \\[1.2ex]
0, & i = j.
\end{cases}
\end{align*}

For degenerate singular values, the Jacobian becomes ill-conditioned and training fails. In our training procedure, we utilized a differentiable singular value decomposition based on the Moore-Penrose pseudoinverse \citep{zhang2024differentiablesvdbasedmoorepenrose} to ensure numerical stability in the truncation. We then keep the top $\cakfprojobsdim$ singular vectors in the update step. 

%% file: 04_experiments.tex
\section{Experiments}
\label{sec:experiments}

\subsection{Synthetic Tasks}
\label{sec: dynamics}

% https://github.com/williamgilpin/dysts/blob/master/dysts/data/all_attractors.png for other attractors

A standard benchmark for previous dynamical latent variable models \citep{Zhao2017VLGP, Pandarinath2018LFADS} is the synthetic Lorenz spiking dataset. In this dataset, the neural dynamics are generated from the Lorenz system of nonlinear ordinary differential equations: 
\begin{align*}
  \dot{x} &= \sigma (y - x), \\
  \dot{y} &= x(\rho - z) - y, \\
  \dot{z} &= xy - \beta z,
  \end{align*}
where we choose the standard $(\sigma, \rho, \beta) = (10, 28, \frac{8}{3})$. 
The dynamics are projected to a space of artificial neurons in $\R^{\obsdim}$ with a random linear projection which produces a rate function $\lambda_t$ for a Poisson random vector $y_t \sim \text{Pois} (\lambda_t)$ at each timestep $t \in [T]$ to generate artificial spikes. Evaluation of performance is between the model predicted firing rates and the true latent firing rates.  We initialize $C$ latent initial conditions with random normal initial states and $R$ trials per condition for $C\cdot R$ total time series, randomly mixing the conditions to force successful models to learn the effects of input pertubations. In addition to denoising, a central difficulty of this task is that slight perturbations in the initial conditions lead to large changes in dynamics. 

\subsubsection{Computational Scaling}

\begin{figure}
  \centering
  \includegraphics[width=0.4\textwidth]{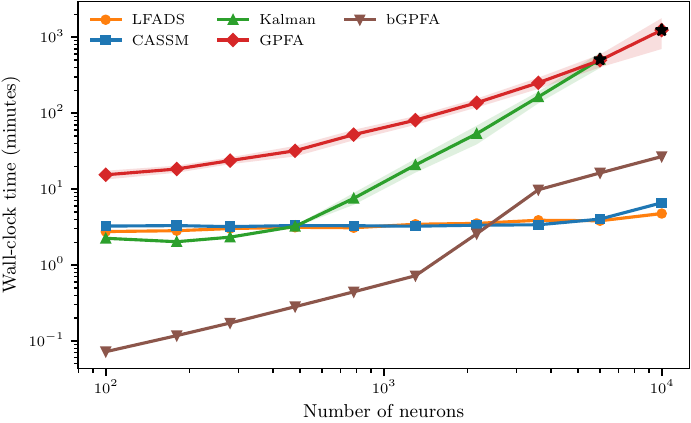}
  \caption{CASSM does not begin to enter its linear scaling regime until around $10^4$ neurons, whereas bGPFA begins its quasilinear regime much earlier. At large neural scales with fixed computational budgets, CASSM is almost a day faster than GPFA or the Kalman Filter, and several hours faster than other scalable Bayesian methods. }
  \label{fig:scaling}
 \end{figure}

\label{sec: computational-scaling}
In Figure \ref{fig:scaling}, we evaluate mean training times across $5$ seed runs for GPFA, bGPFA (an inducing point method that scales GPFA quasilinearly in time instead of cubically), the Kalman Filter trained with a log-likelihood loss, LFADS, and CASSM. The time, in general, represents the completion of $50$ epochs, which was chosen due to its uniformity in reporting and sufficiency for convergence of the majority of methods for most neural scales. We include more justification for our fixed-budget reporting in Appendix \ref{sec:computational-budget}. In Figures \ref{fig:mse100} and \ref{fig:mse10000}, we report $100$ epochs, and as we can see, this cutoff choice is most favorable to GPFA and LFADS, as CASSM typically converges much earlier than $50$ epochs.

Both the Kalman Filter and CASSM are trained with Matérn$\left( \frac{1}{2} \right)$  temporal dynamics models, while GPFA uses a Laplacian kernel, which is a special case of a Matérn$\left( \frac{1}{2} \right)$ kernel. We vary the number of neurons spaced evenly apart in logarithmic space from $10^2$ to $10^4$ neurons, fixing $T=100$ timesteps, $C=10$ conditions, and $R=10$ trials per condition, or $100$ trials. Thus, as the state-space dimensionality increases, the ratio of data to dimensionality is decreasing. 

Our synthetic experiments are designed to vary the balance of trials and neurons. In the scale-imbalanced regime, where the ratio of trials to neurons becomes extremely low, we demonstrate that model bias and the structure of optimization becomes much more important. In log-log space, we find that GPFA scales less favorably in a computational sense than CASSM, LFADS, or bGPFA. At $n=10^4$, the optimization is not able to complete an epoch within a day of computation, which we denote with a black star on the scaling figure (and thus do not report performance for in Figure \ref{fig:mse10000}). 

The Kalman Filter has roughly the same slope as GPFA at larger neural scales, but runs out of memory near $10^4$, and we similarly do not report performance for it in Figure \ref{fig:mse10000}. CASSM and LFADS only begin to reach their linear scaling regimes around $10^4$ neurons, and CASSM is hours faster than bGPFA at this scale. 

We also note memory issues for bGPFA unless the number of Monte Carlo samples is small, even when all methods are only equipped with a latent dimensionality of $3$. CASSM is slightly slower than LFADS, most likely due to the memory overhead of the linear operators used for performing matrix-free computations. This theory is supported by the performance of the Kalman filter, which is actually faster than LFADS for most of the common neural scales ($10^2-10^3$ neurons) of simultaneous recordings due to the efficiency of dense matrix multiplications on GPUs.

\subsubsection{Performance Scaling}

\begin{figure}
  \centering
  \includegraphics[width=0.4\textwidth]{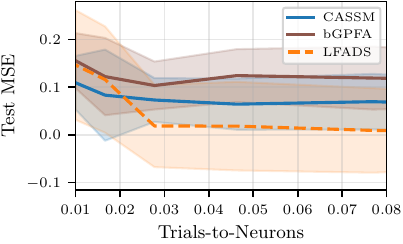}
  \caption{We vary the ratio of trials to neurons, holding trials constant. Here, Bayesian priors can give powerful inductive biases to offset the lack of data. This is evident in the slopes of CASSM (ours) and bGPFA \citep{Jensen2021Scalable}, which are approximate Bayesian methods of the Kalman Filter and Gaussian Process Factor Analysis \citep{Yu2008GPFA}, respectively. When holding the number of trials constant relative to the number of neurons, the Bayesian methods are much less sensitive than LFADS, a nonlinear variational autoencoder \citep{Pandarinath2018LFADS}.}
  \label{fig:snr}
 \end{figure}

\begin{figure}[t]
  \centering
  
  \begin{subfigure}{\columnwidth}
      \centering
      \includegraphics[width=0.7\columnwidth]{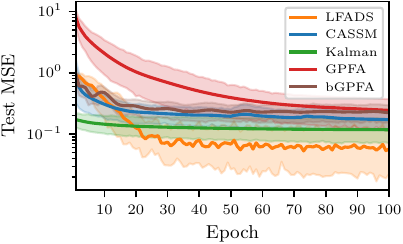}
      \caption{At $10^2$ neurons, where the number of trials equals the number of neurons, LFADS quickly dominates the Bayesian methods in performance. At this scale, all methods train in under 20 minutes, and unless uncertainty is explicitly desired, CASSM's value proposition is minimal.}
      \label{fig:mse100}
  \end{subfigure}
  
  \vspace{6pt}
  
  \begin{subfigure}{\columnwidth}
      \centering
      \includegraphics[width=0.7\columnwidth]{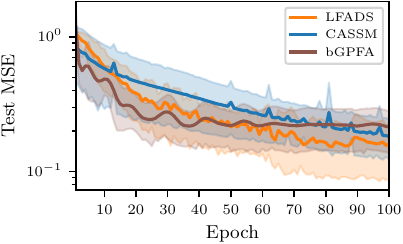}
      \caption{At $10^4$ neurons, the inductive bias of Bayesian methods becomes important when the number of trials is several orders of magnitude smaller than the number of neurons. Given CASSM's favorable computational scaling and performance, its value proposition strengthens even when uncertainty quantification is not explicitly desired.}
      \label{fig:mse10000}
  \end{subfigure}

  \caption{Test MSE across model classes under different neuron-count regimes.}
  \label{fig:mse_combined}
\end{figure}

In Figures \ref{fig:mse100} and \ref{fig:mse10000}, we report MSE for the difference between the true firing rates and the inferred firing rates at various scales from the experiments of Section \ref{sec: computational-scaling}. We notice a trend that holds in real data as well: LFADS dominates when the trials far outnumber the neurons, but degrades rapidly in performance when the trials are limited, a direct result of its flexibility in optimization. Here, both scalable Bayesian methods have significantly more competitive predictive performance, and the separating factor is the meaningfulness of uncertainty and computational or memory complexity.

\subsubsection{Coverage}

For the Bayesian methods, we checked our uncertainty estimates by evaluating the coverage of the $95\%$ confidence intervals for the true firing rates across the same scales as in the previous experiments. The outcomes confirm analysis on SVGPs and GPs from previous papers \citep{Wenger2022PosteriorComputational} \citep{Wang2018batched} in the context of neural data. The results show the need for probabilistic numerical methods in neural data analysis, as the uncertainty quantification of approximation methods is less meaningful when not accounting for computational uncertainty. 

\begin{table}[t]
  \centering
  \footnotesize
  \setlength{\tabcolsep}{3pt}
  \renewcommand{\arraystretch}{1.1}
  \begin{tabular}{lccc}
  \toprule
  & \multicolumn{3}{c}{Coverage ($95\%$)} \\
  & \multicolumn{3}{c}{\scriptsize $n \in \{100,1000,10000\}$} \\
  \cmidrule(lr){2-4}
  Method & $100$ & $1000$ & $10000$ \\
  \midrule
  bGPFA \citep{Jensen2021Scalable}
      & $0.701\!\pm\!0.03$
      & $0.743\!\pm\!0.03$
      & $0.749\!\pm\!0.09$ \\

  GPFA \citep{Yu2008GPFA}
      & $0.893\!\pm\!0.05$
      & $0.878\!\pm\!0.06$
      & N/A \\

  Kalman Filter
      & $0.872\!\pm\!0.03$
      & $0.868\!\pm\!0.03$
      & N/A \\

  CASSM
      & $0.917\!\pm\!0.03$
      & $0.919\!\pm\!0.05$
      & $0.886\!\pm\!0.07$ \\
  \bottomrule
  \end{tabular}
  \caption{Coverage performance for varying orders of magnitude of neurons. Approximation leads to methods that are overconfident in their uncertainty estimates, a problem that CASSM is able to improve upon. We report standard errors across $5$ random seeds.}
  \label{tab:coverage-scale}
\end{table}

\subsection{Real Datasets}
\label{sec: realdatasets}

The Neural Latents Benchmark (NLB) \citep{PeiYe2021NeuralLatents} provides four datasets that span motor, sensory, and cognitive brain regions to evaluate unsupervised latent variable models. The primary scoring tool is a co-smoothing metric (``co-bps'') \citep{Macke2011Empirical} that evaluates the log-likelihood of held-out neurons relative to a control model that predicts the trial-averaged firing rate. There are train/validation/test sets for each of held-in and held-out neurons included by NLB: the goal is not to generalize to unseen neurons, but to infer relationships between neurons that can generalize across stimuli. Held-in and held-out neurons are thus both available at training time. At test time, a held-out neuron subset is predicted given held-in test neurons. Secondary metrics include behavior decoding accuracy and  peristimulus time histogram matching. See Appendix \ref{sec: realdatasets} for more details on the datasets and evaluation metrics.

\subsubsection{ Evaluating Held-Out Neurons}

 For GPFA as well as a Bayesian filter and smoother with a linear dynamics model, conditioning the held-out test neurons on the activity of the held-in test neurons is not directly possible due to latent indeterminacy, as the inferred latent trajectory exists in an arbitrarily rotated and scaled subspace of the true latent manifold via the readout function. These models must be trained on the held-in neurons only, and the map between the held-in dynamics and held-out neurons is learned via a linear decoder after training. 

The model learned by bGPFA can subsequently be used for predictions on held-out data by conditioning
on partial observations as used for cross-validation in \S 3.1 of \citep{Jensen2021Scalable}. In this case, the model is trained on both held-in and held-out neurons.
In the case of a Bayesian filter and smoother with a Matérn dynamics model, as is the case in CASSM, we can directly infer the held-out neurons by training on the entire population and masking the measurement functions for the held-out test neurons during inference.

The results for LFADS and GPFA were tuned by the Neural Latents team \citep{PeiYe2021NeuralLatents} to achieve the best performance on the datasets. We tuned the other methods by a grid search over each method's hyperparameters, and report more training details in the Appendix.

% Requires: \usepackage{booktabs}
\newcommand{\ATwoTrials}{270}
\newcommand{\ATwoNeurons}{20}
\newcommand{\DMFCTrials}{740}
\newcommand{\DMFCNeurons}{50}

\begin{table}[t]
  \centering
  \footnotesize
  \setlength{\tabcolsep}{3pt}
  \renewcommand{\arraystretch}{1.1}
  
  \begin{tabular}{lccccc}
  \toprule
 
  & LFADS 
  & bGPFA 
  & GPFA
  & KF 
  & CASSM \\
  \midrule
  
  \multicolumn{6}{c}{\textbf{Area 2 Bump} \quad 
  \scriptsize Trials: \ATwoTrials \;\; Neurons: \ATwoNeurons} \\
  \cmidrule(lr){1-6}
  
  co-bps     & 0.2569 & 0.1611 & 0.1680 & 0.1699 & 0.1693 \\
  vel $R^2$  & 0.8492 & 0.5573 & 0.5975 & 0.5805 & 0.5818 \\
  psth $R^2$ & 0.6318 & 0.5021 & 0.5289 & 0.5260 & 0.5653 \\
  
  \addlinespace[4pt]
  
  \multicolumn{6}{c}{\textbf{DMFC RSG} \quad 
  \scriptsize Trials: \DMFCTrials \;\; Neurons: \DMFCNeurons} \\
  \cmidrule(lr){1-6}
  
  co-bps     & 0.1829 & 0.1105 & 0.1176 & 0.1119 & 0.1097 \\
  tp corr    & --     & --     & --     & --     & --     \\
  psth $R^2$ & 0.6359 & 0.4657 & 0.2142 & 0.4573 & 0.4333 \\
  
  \bottomrule
  \end{tabular}
  
  \caption{Method performance on NLB real-world datasets.}
  \label{tab:a2b-dmfc-horizontal}
  \end{table}

% Requires: \usepackage{booktabs}
\newcommand{\MCMazeTrials}{1720}
\newcommand{\MCMazeNeurons}{140}
\newcommand{\MCRTTTrials}{810}
\newcommand{\MCRTTNeurons}{100}

\begin{table}[t]
  \centering
  \footnotesize
  \setlength{\tabcolsep}{3pt}
  \renewcommand{\arraystretch}{1.1}
  
  \begin{tabular}{lccccc}
  \toprule

  & LFADS
  & bGPFA
  & GPFA
  & KF
  & CASSM \\
  \midrule
  
  \multicolumn{6}{c}{\textbf{MC Maze} \quad
  \scriptsize Trials: \MCMazeTrials \;\; Neurons: \MCMazeNeurons} \\
  \cmidrule(lr){1-6}
  
  co-bps     & 0.3324 & 0.1840 & 0.1872 & 0.1871 & 0.1859 \\
  vel $R^2$  & 0.9097 & 0.5977 & 0.6399 & 0.6027 & 0.6022 \\
  psth $R^2$ & 0.6360 & 0.3144 & 0.5150 & 0.0548 & 0.1191 \\
  
  \addlinespace[4pt]
  
  \multicolumn{6}{c}{\textbf{MC RTT} \quad
  \scriptsize Trials: \MCRTTTrials \;\; Neurons: \MCRTTNeurons} \\
  \cmidrule(lr){1-6}
  
  co-bps     & 0.1868 & 0.1438 & 0.1548 & 0.1434 & 0.1401 \\
  vel $R^2$  & 0.6167 & 0.4720 & 0.5339 & 0.4552 & 0.4444 \\
  psth $R^2$ & --     & --     & --     & --     & --     \\
  
  \bottomrule
  \end{tabular}
  
  \caption{Method performance on NLB real-world datasets: MC Maze and MC RTT. The real world datasets confirm our previous synthetic experiments, where LFADS dominates when the number of trials is at least the number of neurons. In this regime, the data benefit the flexibility of the LFADS model, and the linear Bayesian methods are unable to capture the complex dynamics.}
  \label{tab:mc-horizontal}
  \end{table}

\subsubsection{Hand Kinematics Visualization}

Figure \ref{fig:velocity} demonstrates the time evolution of behaviorally relevant variables predicted by CASSM: for the Area2 Bump recordings of the somatosensory cortex, these are the hand velocities in the $x-y$ plane for a primate in a center-out reaching task. In a portion of the trials, the monkey's arm was bumped from four different angles with a mechanical pertubation before the reach. Linear decoders fitted on predicted activity yield realistic kinematic outputs with calibrated uncertainty quantification, which is facilitated by sampling from the Gaussian filtering distribution after training and averaging over the linear decoder fits. These results validate the accuracy of the neural activity reconstruction, demonstrating the practical utility of our approach in decoding behaviorally relevant signals.


\subsubsection{Zebrafish Dataset}

Our final experiment considers a large-scale, low-trial dataset of approximately 2.5 million datapoints of fluorescence data from larval zebrafish subjects \citep{Chen2018Zebra}, analyzing roughly 2 orders of magnitude more neurons than the largest experiment in the bGPFA \citep{Jensen2021Scalable} analysis. To enable trial-based learning from recordings, we segmented the continuous hour-long recording by taking advantage of the periodic nature of the experimental stimulus presentations of varying light sources, a procedure which is also used in the trial-averaging cluster analysis of that original paper \citep{Chen2018Zebra} and is common practice in general \citep{Keshtkaran2022AutoLFADS}. We then hold out neurons for evaluation as performed in our Neural Latents Benchmark experiments.

\newcommand{\ZebraTrials}{96}
\newcommand{\ZebraNeurons}{88K}

\begin{table}[t]
  \centering
  \footnotesize
  \setlength{\tabcolsep}{4pt}
  \renewcommand{\arraystretch}{1.05}
  \begin{tabular}{lccc}
      \toprule
      \multicolumn{4}{c}{Zebrafish} \\
      \multicolumn{4}{c}{\scriptsize Trials: \ZebraTrials \quad Neurons: \ZebraNeurons} \\
      \midrule
      Method & MSE & NLL & Time (hrs) \\
      \midrule
      LFADS \citep{Pandarinath2018LFADS} 
          & $0.038 \pm 0.012$ & -- & 3.3 \\

      bGPFA \citep{Jensen2021Scalable} 
          & OOM & OOM & OOM \\

      CASSM-SP
          & $0.035 \pm 0.007$ & $-0.203 \pm 0.078$ & 4.0 \\

      CASSM
          & $0.044 \pm 0.010$ & $-0.132 \pm 0.034$ & 4.3 \\

      \bottomrule
  \end{tabular}
  \caption{Unlike in the Lorenz experiments, where spatial information is absent, the Zebrafish dataset gives approximate $3D$ coordinates for each neuron. CASSM's spatial prior (CASSM-SP) allows it to perform more competitively with LFADS, while also providing calibrated uncertainty quantification at similar computational cost. The performance of bGPFA is not reported due to out-of-memory errors.}
\end{table}

%% file: 05_conclusion.tex
\section{Conclusion}
\label{sec:conclusion}

In this work, we presented CASSM, a probabilistic numerical framework that extends the Computation-Aware Kalman Filter, scales to modern-sized neural datasets, and performs model selection using a novel loss function. In particular, our algorithm is designed for the \emph{scaled-imbalanced regime}, where trials may not scale with the number of recorded neurons and stronger inductive biases are useful for structured learning. We provide a decision tree based on our findings for key dataset properties in Appendix \ref{fig:decision-tree}. We found that at smaller neural dataset sizes with many trials like \citep{PeiYe2021NeuralLatents}, nonlinear methods are clear winners for prediction, while Bayesian methods are essentially ``take-your-pick'' as approximation is not as essential. However, when scaling to larger datasets with fewer relative trials, CASSM suddenly becomes more competitive with nonlinear, deep learning methods that do not model uncertainty. In particular, when exploiting spatial priors in the zebrafish dataset, we found that CASSM and LFADS perform statistically similar. 

\subsection{Limitations}

Tuning CASSM requires choosing the appropriate number of actions, and the rank of the approximate posterior update is not always clear from the problem structure. Furthermore, as a linear method, choosing too few basis functions can severely degrade performance. Several updated versions of GPFA have incorporated more flexible likelihoods and priors into their analyses, while CASSM is limited to GP regression with a conjugate observational noise model. Lastly, we do not address model misspecification, which is not obvious to incorporate into the LGSSM framework. 

\subsection{Future Work}

CASSM can be applied to many different types of spatiotemporal data beyond neuroscience. In addition, our GPU-accelerated algorithm probably has low-hanging fruit to reduce memory overhead of the linear operators and speed up inference. Additional latent dynamical structures are also possible. 

%% file: 98_appendix.tex
\appendix

\numberwithin{figure}{section}
\numberwithin{table}{section}
% \numberwithin{algorithm}{section}

\numberwithin{definition}{section}
\numberwithin{theorem}{section}
\numberwithin{proposition}{section}
\numberwithin{lemma}{section}
\numberwithin{corollary}{section}
\numberwithin{remark}{section}

% Replace all Rs, with Lambdas \obsnoisecov
% replace all H's with bold H's \obsH
% all Q's are the same
% V's are denoted as S \cakfacts
% S is G \ksstate
% don't forget the hats! 

%G_hat: \cakfprojgram
% H_hat: \cakfprojH
% R_hat: \cakfprojobsnoisecov (lambda hat)

\section*{Supplementary Material}

The supplementary materials contain derivations for our theoretical framework and proofs for the mathematical statements in the main text.

We also provide implementation specifics and describe our experimental setup in more detail.

% TOC appendix
\startcontents[sections]
\vspace{1em}
\printcontents[sections]{l}{1}{\setcounter{tocdepth}{3}}
\vspace{3em}

\newpage

\section{Description of the Computational Problem}
\label{sec: problem-description}

\subsection{Experimental Setup and Data Collection}

We study the problem of inferring latent neural dynamics from multi-electrode array recordings collected during controlled behavioural experiments.

\begin{figure*}[h]
  \centering
  \includegraphics[width=\textwidth]{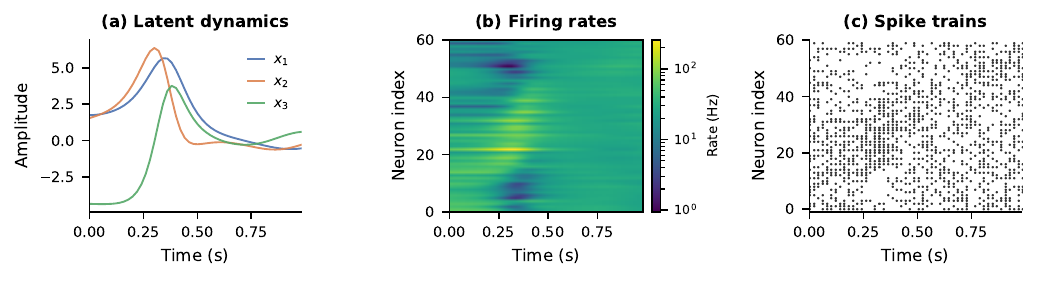}
  \caption{Generative model for neural population data illustrated on the Lorenz system.
  \textbf{(a)}~The three-dimensional latent state $\gmp_k$ evolves according to chaotic Lorenz dynamics; each trace corresponds to one latent coordinate.
  \textbf{(b)}~A linear readout of the latent state passed through a softplus nonlinearity gives per-neuron firing rates (Hz, log scale). Neurons are sorted by their peak-rate time for visual clarity.
  \textbf{(c)}~Spike trains are sampled as Bernoulli draws from the instantaneous firing rates.}
  \label{fig:lorenz-generative}
\end{figure*}

Figure~\ref{fig:lorenz-generative} illustrates the generative structure of such data: a low-dimensional latent trajectory (panel~a) drives per-neuron firing rates (panel~b), from which discrete spike trains are observed (panel~c).

In practice, the latent trajectory is unobserved and must be inferred from spike trains alone.

\subsubsection{Recording technology}
The main datasets used in our real experiments are drawn from the Neural Latents Benchmark (NLB) \citep{PeiYe2021NeuralLatents}, a curated collection of primate electrophysiology recordings.

Neural activity is recorded via surgically implanted microelectrode arrays (high-density silicon probes) targeting specific cortical areas.

Each electrode captures the spiking activity of one or more nearby neurons at sub-millisecond (ms) temporal resolution.

Individual neuron action potentials (``spikes'') are identified through spike-sorting or thresholding and binned into 5\,ms or 20\,ms windows, yielding an integer-valued spike count for each neuron at each time step.

Beyond primate electrophysiology, recent large-scale neural recording efforts have also explored whole-brain imaging in small vertebrates such as the zebrafish.

In particular, the experiments of \citep{Chen2018Zebra} use light-sheet calcium imaging to record activity from tens of thousands of neurons simultaneously across the entire zebrafish brain.

Larval zebrafish are head-fixed while exposed to different frequencies and patterns of light.

The resulting data consist of continuous fluorescence traces with high spatial coverage but lower temporal resolution than electrophysiology.

These datasets provide a complementary regime to spiking recordings: rather than sparse, high-temporal-resolution point processes, they yield dense, smooth signals capturing brain-wide population dynamics, making them particularly well-suited for studying large-scale latent structure and global dynamical patterns.

\subsection{Modeling and State-Space Representation}

In the neural context, the latent state $\gmp_k \in \mathbb{R}^d$ represents a compressed description of the instantaneous \emph{population activity vector} at time step $k$.

The dynamics matrix $\gmpA_k$ encodes how the population state propagates forward in time, capturing recurrent structure such as oscillations, fixed points, or slow drift.

In CASSM, this is parameterised as a discretised linear time-invariant SDE whose continuous-time limit corresponds to a Mat\'ern Gaussian process prior (see Appendix~\ref{sec:lti-sde}), providing a principled way to encode smoothness assumptions over the temporal trajectory.

The process noise $\gmpnoise_k$ accounts for unexplained variability in the latent trajectory (e.g.\ trial-to-trial fluctuations), while $\obsnoise_k$ captures spiking noise and residual variability not explained by the latent representation.

All parameters $\Theta = \{\gmpA_k, \gmpb_k, \gmpnoisecov_k, \obsH_k, \obsnoisecov_k\}_{k=1}^{T}$ are unknown and must be learned from data.

\section{Filtering}
\label{filtering}

\subsection{Objects of Interest}
\label{listobj}

\subsubsection{State-Space Models}

$\gmpA$: Dynamics function

$\gmpnoisecov$: Process noise covariance

$\gmpcov$: Prior (unconditional) state covariance 

$\obsH$: Measurement function

$\obsnoisecov$: Observation noise covariance

$\Theta$: State-space model parameters

\subsubsection{Kalman Filter}

$\kfpmean$: Pre-update (predictive) distribution mean

$\kfpcov$: Pre-update (predictive) distribution variance

$\kfmean$: Filtering distribution mean

$\kfcov$: Filtering distribution variance

$\kfgain$: Kalman gain

$\kfgram$: Innovation matrix

\subsubsection{CASSM}

$\cakfacts$: Policy or action

$\cakfprojH$: Projected measurement function

$\cakfprojobsnoisecov$: Projected observation noise covariance

$\cakfpmean$: Pre-update (predictive) distribution approximate mean

$\cakfpcov$: Pre-update (predictive) distribution approximate covariance

$\cakfmean$: Filtering distribution approximate mean

$\cakfcov$: Filtering distribution approximate covariance

$\cakfprojgram$: Projected innovation matrix

$\cakfpdd$: Pre-update (predictive) belief covariance downdate

$\cakfprojobs$: Projected observation/input

$\cakfw$: Dynamics message mean

$\cakfW$: Dynamics message covariance

$\cakfdd$: Filtering belief covariance downdate

\subsection{Algorithms}
\begin{figure}[h]
  \begin{minipage}[t]{0.56\textwidth}%
    \vspace{-1.7em}
    \begin{algorithm}[H]
      \caption{Computation-Aware State-Space Model (CASSM)} \label{alg:mfkf}
      \begin{algorithmic}[0]
        \small
        \Function{Filter}{$\cakfmean_0, \{ \gmpcov_\idxdtime, \Theta_k, \obs_\idxdtime \}_{\idxdtime = 1}^{n_t}$}
          \State $\cakftdd_0 \gets \mathbf 0 \in \R^{\gmpdim \times r}$ \Comment{Zero matrix}
          \For{$\idxdtime = 1, \dotsc, n_t$}
            \State $\just{\cakfpdd_\idxdtime}{\cakfpmean_\idxdtime} \gets \evallinop{\gmpA_{\idxdtime - 1}}{\cakfmean_{\idxdtime - 1}} + \gmpb_{\idxdtime - 1}$ \Comment{Predict}
            \State $\cakfpdd_\idxdtime \gets \evallinop{\gmpA_{\idxdtime - 1}}{\cakftdd_{\idxdtime - 1}}$
            \State \(
              \cakfmean_\idxdtime, \cakfdd_\idxdtime
              \gets
              \Call{Update}{
                \cakfpmean_\idxdtime, \cakfpdd_\idxdtime,
                \dotsc % \gmpcov_\idxdtime, \obsH_\idxdtime, \obsnoisecov_\idxdtime, \obs_\idxdtime
              }
              \)
            \State $\cakftdd_\idxdtime \gets \Call{PseudoInvSVD}{\cakfdd_\idxdtime}$ \Comment{\citep{zhang2024differentiablesvdbasedmoorepenrose}}
            \State $\mathcal{L}^{(\idxdtime -1)}(\Theta_k) \mathrel{+}= \mathcal{L}^{(\idxdtime)}(\Theta_k)$ \Comment{Eq. \ref{def:loss-function}}
          \EndFor
          \State \textbf{return} \( \{ \cakfmean_\idxdtime, \cakfdd_\idxdtime \}_{\idxdtime = 1}^{n_t} \)
        \EndFunction
      \end{algorithmic}
    \end{algorithm}
  \end{minipage}
  \hfill
  \begin{minipage}[t]{0.42\textwidth}%
    \vspace{-1.7em}
    \begin{algorithm}[H]
      \caption{CASSM Update Step} \label{alg:projected_update}
      \begin{algorithmic}[0]
        \small
        \vspace{-.55ex}
        \Function{Update}{$\cakfpmean, \cakfpdd, \gmpcov, \obsH, \obsnoisecov, \obs$}
          \State \textcolor{blue}{$\just{\cakfprojH\T}{\cakfpcov} \gets \gmpcov - \cakfpdd (\cakfpdd)\T$}
          \State $\just{\cakfprojH\T}{\textcolor{blue}{\cakfacts}} \gets \Call{Policy}{\cakfpmean, \cakfpcov, \dotsc}$
          \State $\just{\cakfprojH\T}{\cakfprojH\T} \gets \obsH\T \textcolor{blue}{\cakfacts}$
          \State $\just{\cakfprojH\T}{\cakfprojobsnoisecov} \gets \textcolor{blue}{\cakfacts}\T \textcolor{blue}{\obsnoisecov}\textcolor{blue}{\cakfacts}$
          \State $\just{\cakfprojH\T}{\cakfprojobs} \gets \textcolor{blue}{\cakfacts}\T \obs$
          \State $\just{\cakfprojH\T}{\cakfprojgram} \gets \cakfprojH \textcolor{blue}{\cakfpcov} \cakfprojH\T + \cakfprojobsnoisecov$
          \State $\just{\cakfprojH\T}{\cakfw} \gets \cakfprojH\T \cakfprojgram\pinv (\cakfprojobs - \cakfprojH \cakfpmean)$
          \State $\just{\cakfprojH\T}{\cakfW} \gets \cakfprojH\T (\cakfprojgram\pinv)^\frac{1}{2}$
          % \Comment{$(\cakfprojgram\pinv)^\frac{1}{2} (\cakfprojgram\pinv)^\frac{\top}{2} = \cakfprojgram\pinv$}
          \State $\just{\cakfprojH\T}{\cakfmean} \gets \cakfpmean + \textcolor{blue}{\cakfpcov} \cakfw$
          \State \(
          \just{\cakfprojH\T}{\cakfdd}
          \gets
          \begin{pmatrix}
            \cakfpdd & \textcolor{blue}{\cakfpcov} \cakfW
          \end{pmatrix}
          \)
          \State \textbf{return} $(\cakfmean, \cakfdd)$
        \EndFunction
      \end{algorithmic}
    \end{algorithm}
  \end{minipage}
  \vspace{-1em}
\end{figure}

Similar to how the square-root Kalman Filter implements the Kalman Filter, Algorithms \ref{alg:mfkf} and \ref{alg:projected_update} mathematically implement the Computation-Aware Kalman Filter (CAKF) \citep{Pfoertner2024ComputationAware}, but require modifications to be computationally efficient and differentiable. 

Most notably, the truncation step in \citep{Pfoertner2024ComputationAware} cannot be stably differentiated through, so we replace this with a pseudo-inverse of the covariance downdate as discussed in \ref{sec: procedure}. 

We highlight matrix-free objects in Algorithms \ref{alg:mfkf} and \ref{alg:projected_update} in blue. 

The pre-update latent covariance $\cakfpcov$ is already the difference between a Kronecker-product linear operator and a root linear operator, so it admits efficient lazy multiplication without explicitly materializing a dense covariance. 

In CASSM, the policy $\cakfacts$ introduces an additional challenge: it is itself represented as a matrix-free object, implemented as a \texttt{BlockDiagonalSparseLinearOperator} in the \texttt{LinearOperator} Python package.

This creates several design constraints that do not arise in a dense implementation. 

First, model selection requires evaluating loss terms that depend not only on filtered means, but also on covariance-derived quantities such as projected noise terms, traces, diagonals, and log-determinants. 

When both $\cakfpcov$ and $\cakfacts$ are lazy operators, one must decide which of these quantities can be computed through matrix-vector products alone, and which require partial materialization. 

For example, the projected observation covariance $\cakfprojobsnoisecov = \cakfacts^\top \obsnoisecov \cakfacts$ and the projected Gram matrix $\cakfprojgram = \cakfprojH \cakfpcov \cakfprojH^\top + \cakfprojobsnoisecov$ may remain implicit during filtering, but loss evaluation requires their diagonals, traces, Cholesky factors, or log-determinants. 

These are not interchangeable computationally: a diagonal is trivial to extract, whereas a log-determinant generally forces some smaller dense representation after projection.

\section{Choosing Between Latent Variable Models}

\subsection{Description of Competing Models}

We compare CASSM against four baseline models. 

\paragraph{GPFA} \citep{Yu2008GPFA}
is a linear-Gaussian latent variable model that jointly performs temporal smoothing
and dimensionality reduction using Gaussian-process priors over latent trajectories.
Inference is exact but requires cubic time in the number of time steps and quadratic
memory, making it impractical for long recordings.

\paragraph{bGPFA} \citep{Jensen2021Scalable}
extends GPFA with a variational objective and Bayesian loading matrix, and uses
structured approximations to reduce the cost of temporal inference. Training scales
approximately linearly in the number of time steps (up to a logarithmic factor), but
also grows linearly with the number of neurons and Monte Carlo samples, and
quadratically with the latent dimensionality.

\paragraph{LFADS} \citep{Pandarinath2018LFADS}
is a sequential variational autoencoder with recurrent neural network dynamics.
It is highly expressive and capable of modeling nonlinear trajectories, but training
requires backpropagation through time and scales linearly in the number of time steps
and neurons, and quadratically in the size of the recurrent hidden states.

\paragraph{Kalman Filter} \citep{Sarkka2023BayesianFiltering}
provides exact recursive inference for linear-Gaussian state-space models. Each update
requires dense matrix operations on both the latent state and observation dimensions,
leading to cubic scaling in both the number of neurons and the latent dimensionality,
with quadratic memory requirements.

\subsection{Tradeoffs and Decision Analysis}
\begin{figure*} 
  \centering
  \includegraphics[width=\textwidth]{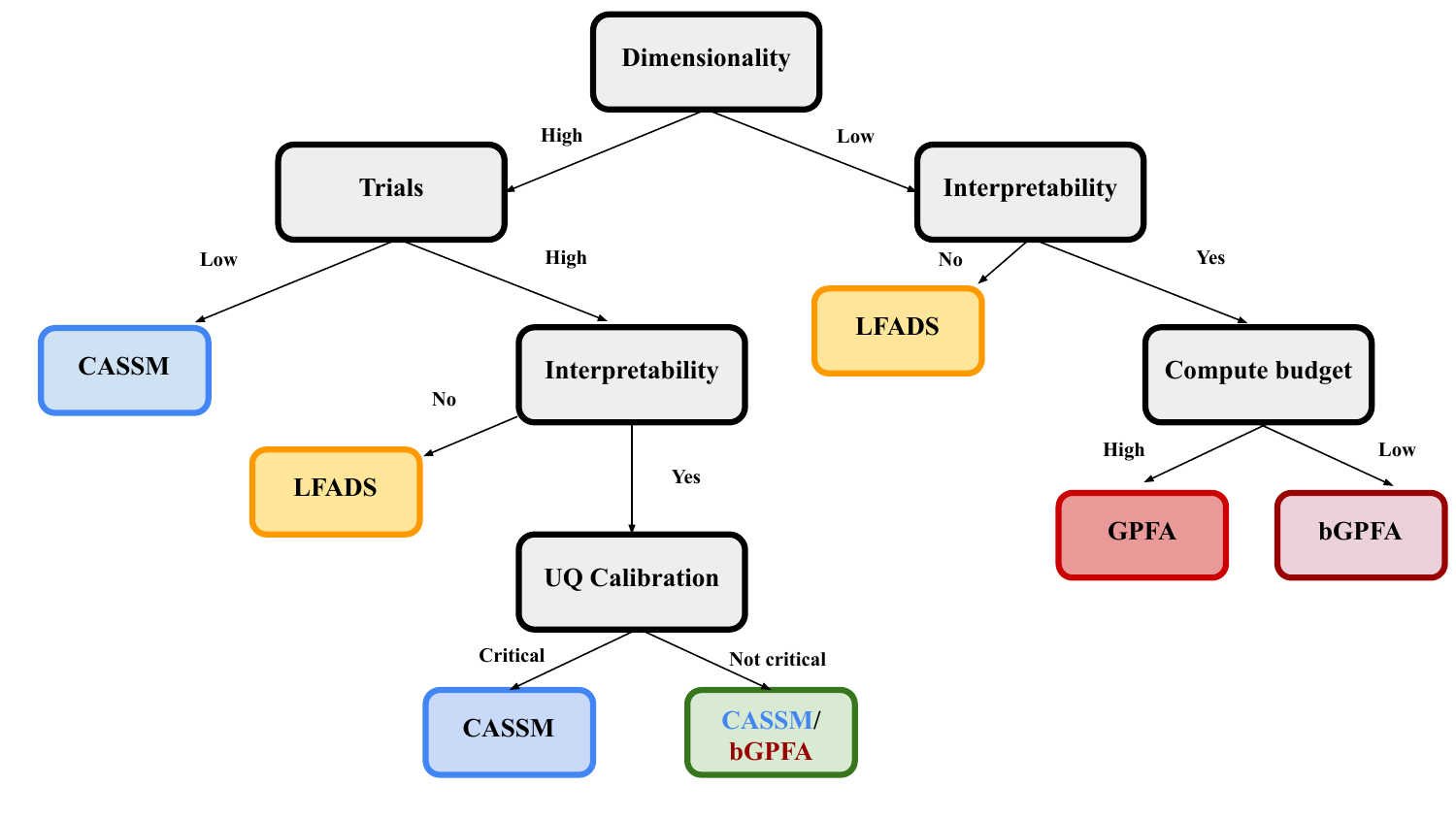}
  \caption{Decision analysis for choosing between latent variable models based on our experimental findings.}
  \label{fig:decision-tree}
\end{figure*}

Because of the variety of models and algorithms for this problem, an important contribution of our analysis is to provide a basic decision tree for choosing between latent variable models in neuroscience in Figure \ref{fig:decision-tree}.

We demonstrated that in the scaled-imbalanced regime, CASSM has superior mean predictive performance to LFADS in the real and synthetic experiments. 

However, LFADS is a more flexible model class and can be expected to outperform CASSM in the low scale-imbalanced regime. 

In this case, the performance gap is substantial and the justification to use CASSM is quite low.

We were surprised to see a basic Kalman filter with model selection to perform so well at these smaller scales, often more accurate and faster than GPFA. 

We note that naive implementation of the Kalman Filter, though not included in our final experiments, can be very difficult to optimize. 

This is the simple linear dynamics and linear observation model, where one learns these matrices directly with no structure. 

We added manifold parameterizations (e.g. Stiefel manifold, etc.) to the dynamics and observation matrices, which substantially improved optimization, but still encountered difficulties with proper initialization and training stability. 

As we explore in Appendix \ref{sec:lti-sde}, the kernel formulation of the Kalman Filter is mathematically equivalent to the Gaussian Process formulation, but is much easier to optimize in practice.

\section{Deriving the Objective Function}
\label{sec:loss-function-derivation}
\begin{definition}[CASSM Objective ]
  \label{def:loss-function}
  
  The CASSM objective function is: 
  \begin{equation*}
    \begin{aligned}
        \mathcal{L} (\Theta) &= -\log p \left( \obsrv_{1:\idxdtimefinal} \mid \Theta \right) + \sum_{k \in [K]} D_{KL} \left(q (\gmp_\idxdtime \mid \obsrv_{1 : \idxdtime}) \parallel p (\gmp_\idxdtime \mid \obsrv_{1 : \idxdtime}) \right)
    \end{aligned}
  \end{equation*}

\end{definition}

  Our primary objects of interest are the pre-update Kalman predictive distribution:

\begin{equation*}
  p (\gmp_\idxdtime \mid \obsrv_{1 : \idxdtime -1 }) \sim \mathcal{N} \left(\gmp_\idxdtime ;  \kfpmean_\idxdtime, \kfpcov_\idxdtime\right)
  \end{equation*}

  and the approximate filter distribution:
  \begin{equation*}
    q (\gmp_\idxdtime \mid \obsrv_{1 : \idxdtime}) \sim \mathcal{N} \left(\gmp_\idxdtime ; \cakfmean_\idxdtime, \cakfcov_\idxdtime\right)
  \end{equation*}

  The motivation of $\mathcal{L}$ is to directly regularize each term of the log-marginal likelihood with the KL-divergence of the compressed filter and the original filter. 
  
  Because the log-likelihood is a linear function of $\Theta$, this reduces to a recursion in the style of \citep{Sarkka2023BayesianFiltering}:

\begin{equation*}
    \mathcal{L}^{(\idxdtime)}(\Theta) = \mathcal{L}^{(\idxdtime -1)}(\Theta) -\log p\left(\obsrv_\idxdtime \mid \obsrv_{\idxdtime - 1}\right)+D_{KL} \left(q \left(\gmp_\idxdtime \mid \obsrv_{1:\idxdtime}\right) \parallel p\left(\gmp_\idxdtime \mid \obsrv_{1:\idxdtime}\right)\right) 
\end{equation*}

 We will consider the update $\Delta (\Theta) = \mathcal{L}^{(\idxdtime)}(\Theta) - \mathcal{L}^{(\idxdtime -1)}(\Theta)$. 

 Our goal is to show that the update corresponds to an evidence lower bound from the variational approximation $q$ to the true posterior $p$. 

\begin{equation*}
    \begin{aligned}
      \Delta (\Theta) &=-\log p\left(\obsrv_\idxdtime \mid \obsrv_{\idxdtime - 1}\right)+D_{KL} \left(q \left(\gmp_\idxdtime \mid \obsrv_{1:\idxdtime}\right) \parallel p\left(\gmp_\idxdtime \mid \obsrv_{1:\idxdtime}\right)\right) \\
      & =-\log p\left(\obsrv_\idxdtime \mid \obsrv_{\idxdtime - 1}\right)+\E_{q\left(\gmp_\idxdtime | \obsrv_{1:\idxdtime}\right)}\left[\log \frac{q \left(\gmp_\idxdtime \mid \obsrv_{1:\idxdtime}\right)}{p \left(\gmp_\idxdtime \mid \obsrv_{1:\idxdtime}\right)}\right] \\
      & =-\log p\left(\obsrv_\idxdtime \mid \obsrv_{\idxdtime - 1}\right)+\E_{q} \left[\log q\left(\gmp_\idxdtime \mid \obsrv_{1:\idxdtime}\right)\right] -\E_{q}\left[\log p\left(\gmp_\idxdtime \mid  \obsrv_{1:\idxdtime}\right)\right] \\
      & =-\log p\left(\obsrv_\idxdtime \mid \obsrv_{\idxdtime - 1}\right)+\E_{q}\left[\log q\left(\gmp_\idxdtime \mid \obsrv_{1:\idxdtime}\right)\right]-\E_{q}\left[\log \frac{p\left(\gmp_\idxdtime, \obsrv_\idxdtime \mid \obsrv_{\idxdtime - 1}\right)}{p\left(\obsrv_\idxdtime \mid \obsrv_{1:\idxdtime-1}\right)}\right] \\
      & =-\log p\left(\obsrv_\idxdtime \mid \obsrv_{\idxdtime - 1}\right)+\E_{q}\left[\log q\left(\gmp_\idxdtime \mid \obsrv_{1:\idxdtime}\right)\right]-\E_{q}\left[\log p\left(\gmp_\idxdtime, \obsrv_\idxdtime \mid \obsrv_{\idxdtime - 1}\right)\right] +\E_{q}\left[\log p\left(\obsrv_\idxdtime \mid \obsrv_{\idxdtime - 1}\right)\right] \\
      & =\E_{q}\left[\log q\left(\gmp_\idxdtime \mid \obsrv_{1:\idxdtime}\right)\right]-\E_{q}\left[\log p\left(\gmp_\idxdtime, \obsrv_\idxdtime \mid \obsrv_{\idxdtime - 1}\right)\right] \\
      & =\E_{q}\left[\log q\left(\gmp_\idxdtime \mid \obsrv_{1:\idxdtime}\right)\right]-\E_{q}\left[\log \left(p\left(\obsrv_\idxdtime \mid \gmp_\idxdtime, \obsrv_{\idxdtime - 1}\right) \cdot p\left(\gmp_\idxdtime \mid \obsrv_{\idxdtime - 1}\right)\right)\right] \\
      & = \E_{q}\left[\log q\left(\gmp_\idxdtime \mid \obsrv_{1:\idxdtime}\right)\right] - \E_{q}\left[\operatorname {log} p\left(\obsrv_\idxdtime \mid \gmp_\idxdtime, \obsrv_{\idxdtime - 1}\right)\right] - \E_{q}\left[\log p\left(\gmp_\idxdtime \mid \obsrv_{\idxdtime - 1}\right) \right]\\
      & =\E_{q}\left[\log p (\obsrv_\idxdtime \mid \gmp_\idxdtime ) \right] + D_{KL} \left( q(\gmp_\idxdtime \mid \obsrv_{1:\idxdtime} ) \parallel p(\gmp_\idxdtime \mid \obsrv_{\idxdtime - 1}) \right)
      \end{aligned}
  \end{equation*}

  We now examine the expectation and KL-divergence terms, respectively. 

  \subsection{Expectation Term}
  \label{expterm}
  
  \begin{equation*}
    \begin{aligned}
      \E_{q\left(\gmp_\idxdtime \mid \obsrv_{1:\idxdtime}\right)} \left[\log p(\obsrv_\idxdtime \mid \gmp_\idxdtime) \right] &= \mathbb{E}_{q(\gmp_\idxdtime \mid \obsrv_{1:\idxdtime})} \left[\log \mathcal{N} (\obsrv_\idxdtime; \obsH \gmp_\idxdtime, 
      \obsnoisecov)\right] \\
      &= \E_{q\left(\gmp_\idxdtime \mid \obsrv_{1:\idxdtime}\right)} \left[ \frac{1}{2} \log \det (\obsnoisecov) + \frac{1}{2} (\obsrv_\idxdtime - \obsH \gmp_\idxdtime)^T \obsnoisecov^{-1}(\obsrv_\idxdtime - \obsH \gmp_\idxdtime) + \frac{1}{2}\obsdim \log(2\pi)\right] \\
      &= \frac{1}{2} \left[\log \det (\obsnoisecov) + \mathbb{E}_q \left[ (\obsrv_\idxdtime - \obsH \gmp_\idxdtime)^T \obsnoisecov^{-1}(\obsrv_\idxdtime - \obsH \gmp_\idxdtime) \right] + \obsdim \log (2\pi) \right] \\
      &= \frac{1}{2} \left[ \log \det (\obsnoisecov) + (\obsrv_\idxdtime - \obsH \cakfmean_\idxdtime)^T \obsnoisecov^{-1} (\obsrv_\idxdtime - \obsH \cakfmean_\idxdtime) +  \operatorname{tr}(\obsnoisecov^{-1} \obsH\cakfcov_\idxdtime \obsH\T) + \obsdim \log (2\pi) \right]\\
      % Trace term compares the (predictive) variance for each neuron to the observation noise (division per neuron) and then sums up the result
      %NOTE: Assumes observation noise covariance is diagonal
      %&=\frac{1}{2} \left[ \log \det (\obsnoisecov) + (\obsrv_\idxdtime - \obsH \cakfmean_\idxdtime)^T \obsnoisecov^{-1} (\obsrv_\idxdtime - \obsH \cakfmean_\idxdtime) +  \sum_\obsdim \frac{\operatorname{diag}(\obsH\cakfcov_\idxdtime \obsH\T)_\obsdim}{\obsnoisecov_{\obsdim, \obsdim}} + \obsdim \log (2\pi) \right]\\
    %   &=\frac{1}{2} \left[ \log \det (\obsnoisecov) + (\obsrv_\idxdtime - \obsH \cakfmean_\idxdtime)^T \obsnoisecov^{-1} (\obsrv_\idxdtime - \obsH \cakfmean_\idxdtime) +  \operatorname{tr}(\obsnoisecov^{-1} \obsH(\gmpcov_\idxdtime - \cakfdd_\idxdtime\cakfdd_\idxdtime\T)\obsH\T) + \obsdim \log (2\pi) \right]\\
    %   &=\frac{1}{2} \left[ \log \det (\obsnoisecov) + (\obsrv_\idxdtime - \obsH \cakfmean_\idxdtime)^T \obsnoisecov^{-1} (\obsrv_\idxdtime - \obsH \cakfmean_\idxdtime) +  \operatorname{tr}(\obsnoisecov^{-1} \obsH\gmpcov_\idxdtime\obsH\T) - \operatorname{tr}(\obsnoisecov^{-1}\obsH\cakfdd_\idxdtime\cakfdd_\idxdtime\T\obsH\T) + \obsdim \log (2\pi) \right]\\
    % &=\frac{1}{2} \left[ \log \det (\obsnoisecov) + (\obsrv_\idxdtime - \obsH \cakfmean_\idxdtime)^T \obsnoisecov^{-1} (\obsrv_\idxdtime - \obsH \cakfmean_\idxdtime) +  \operatorname{tr}(\obsnoisecov^{-1} \obsH\gmpcov_\idxdtime\obsH\T) - \operatorname{tr}(\cakfdd_\idxdtime\T\obsH\T\obsnoisecov^{-1}\obsH\cakfdd_\idxdtime) + \obsdim \log (2\pi) \right]
    \end{aligned}
  \end{equation*}

  \subsection{KL Term}
  \label{klterm}

  \begin{equation*}
    \begin{aligned}
      D_{KL}\left(q\left(\gmp_\idxdtime \mid \obsrv_{1:\idxdtime}\right) \parallel p\left(\gmp_\idxdtime \mid \obsrv_{1:\idxdtime-1}\right)\right) &= D_{KL} \left( \mathcal{N} (\gmp_\idxdtime; \cakfmean_\idxdtime, \cakfcov_\idxdtime) \parallel \mathcal{N} (\gmp_\idxdtime; \kfpmean_\idxdtime, \kfpcov_\idxdtime)  \right) \\
      &= \frac{1}{2} \left[  \operatorname{tr}(\kfpcov_\idxdtime \cakfcov_\idxdtime) -\obsdim + (\kfpmean_\idxdtime - \cakfmean_\idxdtime)^T (\kfpcov_\idxdtime)^{-1}(\kfpmean_\idxdtime - \cakfmean_\idxdtime) + \log \det (\kfpcov_\idxdtime) \right.\\
      &- \left. \log \det (\cakfcov_\idxdtime)\right] \\
      &= \frac{1}{2} \left[  \operatorname{tr}((\kfpcov_\idxdtime)^{-1} (\underbrace{\kfpcov_\idxdtime - \kfpcov_\idxdtime \cakfw_\idxdtime \cakfw_\idxdtime^T \kfpcov_\idxdtime}_{\cakfcov_\idxdtime})) - \obsdim \right. \\ 
      &+ \left. (\kfpmean_\idxdtime - (\underbrace{\kfpmean_\idxdtime + \kfpcov_\idxdtime \cakfw_\idxdtime}_{\cakfmean_\idxdtime}))^T (\kfpcov_\idxdtime)^{-1} (\kfpmean_\idxdtime - \kfpmean_\idxdtime - \kfpcov_\idxdtime \cakfw_\idxdtime) \right.\\
      & \left. - \log \det ((\kfpcov_\idxdtime)^{-1} \cakfcov_\idxdtime )\right] \\
      &= \frac{1}{2} \left[  \operatorname{tr}(\mI_{\obsdim \times \obsdim} - \cakfw_\idxdtime \cakfw_\idxdtime^T \kfpcov_\idxdtime) - \obsdim + \cakfw_\idxdtime^T \kfpcov_\idxdtime \cakfw_\idxdtime - \log \det (\mI_{\obsdim \times \obsdim} - \cakfw_\idxdtime \cakfw_\idxdtime^T \kfpcov_\idxdtime)\right] \\
      &= \frac{1}{2} \left[ \obsdim -  \operatorname{tr}(\cakfw_\idxdtime \cakfw_\idxdtime^T \kfpcov_\idxdtime ) - \obsdim + \cakfw_\idxdtime^T \kfpcov_\idxdtime \cakfw_\idxdtime - \log \det (\mI_{\obsdim \times \obsdim} - \underbrace{\cakfw_\idxdtime \cakfw_\idxdtime^T \kfpcov_\idxdtime}_{\cakfprojH^T \cakfprojgram_{\idxdtime}^{-1} \cakfprojH \kfpcov_\idxdtime})\right] \\
  \end{aligned}
  \end{equation*}

  We are able to work in $\R^{\cakfprojobsdim}$ instead of $\R^{\obsdim}$ in the log-determinant term due to the matrix determinant lemma:
  
  \begin{equation*}
    \begin{aligned}
      \det (\mI + \mU \mV^T) &= \det (\mI + \mV^T \mA^{-1} \mU) \cdot \det (\mA)
    \end{aligned}
  \end{equation*}
  
  Continuing with the expansion:

  \begin{equation*}
    \begin{aligned}
      &= \frac{1}{2} \left[ \cakfw_\idxdtime^T \kfpcov_\idxdtime \cakfw_\idxdtime -  \operatorname{tr} (\cakfprojgram_{\idxdtime}^{-1} \cakfprojH \kfpcov_\idxdtime \cakfprojH^T) - \log \det (\mI_{\cakfprojobsdim \times \cakfprojobsdim} - \cakfprojgram_{\idxdtime}^{-1} \cakfprojH^T \kfpcov_\idxdtime \cakfprojH)\right] \\
      &= \frac{1}{2} \left[\cakfw_\idxdtime^T \kfpcov_\idxdtime \cakfw_\idxdtime -  \operatorname{tr}(\cakfprojgram_{\idxdtime}^{-1} \cakfprojH \kfpcov_\idxdtime \cakfprojH^T) - \log \det (\cakfprojgram_{\idxdtime}^{-1} (\underbrace{\cakfprojgram_{\idxdtime} - \cakfprojH^T \kfpcov_\idxdtime \cakfprojH}_{\cakfacts^T \obsnoisecov \cakfacts})) \right] \\
      &= \frac{1}{2} \left[\cakfw_\idxdtime^T \kfpcov_\idxdtime \cakfw_\idxdtime -  \operatorname{tr}(\cakfprojgram_{\idxdtime}^{-1} \cakfprojH \kfpcov_\idxdtime \cakfprojH^T) - \log \det (\cakfprojgram_{\idxdtime}^{-1} \cakfacts^T \obsnoisecov \cakfacts)  \right]\\
      &= \frac{1}{2} \left[\cakfw_\idxdtime^T \kfpcov_\idxdtime \cakfw_\idxdtime - \operatorname{tr}(\mI_{\cakfprojobsdim \times \cakfprojobsdim}) +   \operatorname{tr}(\cakfprojgram_{\idxdtime}^{-1} \cakfacts^T \obsnoisecov \cakfacts) - \log \det (\cakfprojgram_{\idxdtime}^{-1} \cakfacts^T \obsnoisecov \cakfacts)  \right]
    \end{aligned}
  \end{equation*}

  \subsection{Final numerical form}
  \label{numerical}

 For timestep $\idxdtime$, combining the results of \ref{expterm} and \ref{klterm} we obtain the final numerical form of Definition \ref{def:loss-function}:

  \begin{lemma}
    \label{formnumeric}
    \begin{equation*}
      \begin{aligned}
        \mathcal{L}^{(\idxdtime)}(\Theta) &= \frac{1}{2} \left[ \log \det (\obsnoisecov) + (\obsrv_\idxdtime - \obsH \cakfmean_\idxdtime)^T \obsnoisecov^{-1} (\obsrv_\idxdtime - \obsH \cakfmean_\idxdtime) +  \operatorname{tr}(\obsnoisecov^{-1} \obsH \cakfcov_\idxdtime \obsH\T) + \obsdim \log (2\pi) \right]\\ &+ \frac{1}{2} \left[\cakfw_\idxdtime^T \kfpcov_\idxdtime \cakfw_\idxdtime - \cakfprojobsdim +   \operatorname{tr}(\cakfprojgram_{\idxdtime}^{-1} \cakfacts^T \obsnoisecov \cakfacts) - \log \det (\cakfprojgram_{\idxdtime}^{-1} \cakfacts^T \obsnoisecov \cakfacts)  \right]
      \end{aligned}
    \end{equation*}
  \end{lemma}

  All quantities of the CASSM loss are inexpensive to compute. 
  
  The most concerning term is the inverse of the measurement noise function, $\obsnoisecov$. As a dense matrix, this operation requires $O(\obsdim^3)$ time. 
  
  Experimentally, however, such a measurement device is unlikely to exist in practice: channels can only capture so much long-range dependency (hence the need for hundreds of channels for one probe), and any noise structure can be reasonably treated as at most block-diagonal with a uniformly bounded block size. 
  
  Therefore, its inverse can be computed in $O(\obsdim)$ time. 
  
  The other inverse, $\cakfprojgram_{\idxdtime}^{-1}$, exists in $\R^{\cakfprojobsdim \times \cakfprojobsdim}$. Here, the projection solves the unfavorable scaling from (P1), as $\cakfprojobsdim << \obsdim $.

\section{Connection to PCA}
\label{pca}

\textbf{Entropy in Time and the Greedy Policy}

The entropy of a Gaussian random vector $\mathbf{x} \in \R^n \sim \gaussian{\gmpmean}{\gmpcov}$ is given by:
\begin{equation*}
  \begin{aligned}
    \mathrm{H}(\mathbf{x}) = \frac{n}{2} (1 + \log (2\pi)) + \frac{1}{2} \log \det \gmpcov
  \end{aligned}
\end{equation*}
Now, let the total latent reduction in entropy be defined as:
\begin{equation*}
  \begin{aligned}
    \tilde{\mathrm{H}} (\mathbf{x}) := \mathrm{H}(\mathbf{x}_{1:T}) - \mathrm{H}(\mathbf{x}_{1:T} \mid \mathbf{y}_{1:T})
  \end{aligned}
\end{equation*}
where $\mathbf{x}_{1:T}$ is the set of all latent states and $\mathbf{y}_{1:T}$ is the set of all measurements. By the chain rule for entropy, we have:
\begin{equation*}
  \begin{aligned}
    \mathrm{H}(\mathbf{x}_{1:T}) &= \sum_{k=1}^T \mathrm{H} (\mathbf{x}_k \mid \mathbf{x}_{1:k-1}) 
  \end{aligned}
\end{equation*}
Because we assume a linear Gaussian state-space prior, we also have:
\begin{equation*}
  \begin{aligned}
    \mathrm{H} (\mathbf{x}_k \mid \mathbf{x}_{1:k-1}) &= \mathrm{H}(\mathbf{x}_k \mid \mathbf{y}_{1:k-1})
  \end{aligned}
\end{equation*}
which implies
\begin{equation*}
  \begin{aligned}
    \mathrm{H}(\mathbf{x}_{1:T}) &= \sum_{k=1}^T \mathrm{H}(\mathbf{x}_k \mid \mathbf{y}_{1:k-1}) 
  \end{aligned}
\end{equation*}
However, the Gauss-Markov structure also implies:
\begin{equation*}
  \begin{aligned}
    \mathrm{H}(\mathbf{x}_k \mid \mathbf{x}_{1:k-1}, \mathbf{y}_{1:T}) &= \mathrm{H}(\mathbf{x}_k \mid \mathbf{y}_{1:k})
  \end{aligned}
\end{equation*}
because the current latent state only depends on the previous latent state. Taking these two ideas together, we conclude:
\begin{equation*}
  \begin{aligned}
    \mathrm{H}(\mathbf{x}_{1:T} \mid \mathbf{y}_{1:T}) &= \sum_{k=1}^T \mathrm{H}(\mathbf{x}_k \mid \mathbf{y}_{1:k})
  \end{aligned}
\end{equation*}
So the total reduction in entropy is:
\begin{equation*}
  \begin{aligned}
    \tilde{\mathrm{H}} (\mathbf{x}) &= \sum_{k=1}^T \mathrm{H} (\mathbf{x}_k \mid \mathbf{y}_{1:k-1})  - \sum_{k=1}^T \mathrm{H}(\mathbf{x}_k \mid \mathbf{y}_{1:k})\\ 
    &= \sum_{k=1}^T \mathrm{H} (\mathbf{x}_k \mid \mathbf{y}_{1:k-1}) - \mathrm{H}(\mathbf{x}_k \mid \mathbf{y}_{1:k})
  \end{aligned}
\end{equation*}

In other words, the Gaussian entropy for the difference in prior and posterior Kalman distributions decomposes as a sum over time indexes $k \in [T]$. In general:
\begin{align*}
  \max_{\cakfacts_{k \in [T]}} \tilde{\mathrm{H}} (\mathbf{x}) &= \max_{\cakfacts_{k \in [T]}} \sum_{k=1}^{T} \frac{1}{2} \left(\log \det \cakfpcov_\idxdtime - \log \det  \cakfcov_\idxdtime \right)\\ &\leq \sum_{k=1}^{T} \max_{\cakfacts_{k \in [T]}} \frac{1}{2} \left(\log \det \cakfpcov_\idxdtime - \log \det  \cakfcov_\idxdtime \right)
\end{align*}

Therefore, to achieve the upper bound of differential entropy, it suffices to choose a policy $\cakfacts_k$ given a sequence of policies $\cakfacts_1, \ldots, \cakfacts_{k-1}$ for each $k \in [T$]. 

Given that the Kalman Filter is an online algorithm, the greedy policy at time $\idxdtime$ is a natural strategy for this problem.

\textbf{Update Equations and Eigenvalue Maximization}

Note that in the exact Kalman Filter, our posterior covariance update reads:
\begin{equation*}
  \begin{aligned}
    \kfcov_\idxdtime = \kfpcov_\idxdtime - \kfpcov_\idxdtime \obsH_k^T (\obsH_k \kfpcov_\idxdtime \obsH_k^T + \obsnoisecov_k)^{-1} \obsH_\idxdtime \kfpcov_\idxdtime
  \end{aligned}
\end{equation*}
And in CASSM, an approximate Kalman Filter, we restrict $\obsrv$ to the subspace spanned by $\cakfacts$, or:
\begin{equation*}
  \begin{aligned}
    \cakfcov_\idxdtime = \cakfpcov_\idxdtime - \cakfpcov_\idxdtime \obsH_k^T \cakfacts_k (\cakfacts_k^T \obsH_k \cakfpcov_\idxdtime \obsH_k^T \cakfacts_k + \cakfacts_k^T \obsnoisecov_{k} \cakfacts_k)^{-1} \cakfacts_k^T \obsH_k \cakfpcov_\idxdtime
  \end{aligned}
\end{equation*}
We consider the problem of choosing $\cakfacts_k$, or:
\begin{align*}
\argmax_{\cakfacts_k} \mathrm{H}^{(k)} & = \argmax_{\cakfacts_k} \mathrm{H}\left(\cakfpstate_\idxdtime\right)-\mathrm{H}\left(\cakfstate_\idxdtime\right) \\
      &=\argmax_{\cakfacts_k}  \frac{1}{2} \log \frac{\det \cakfpcov_\idxdtime}{\det \cakfcov_\idxdtime}
\end{align*}

Define $\cakfprojgram_{\idxdtime} := \cakfacts_k^{\top}\left(\obsH_{k} \cakfpcov_\idxdtime \obsH_{k}^{\top}+\obsnoisecov_{k}\right) \cakfacts_k =\cakfacts_k^{\top}\kfgram_k \cakfacts_k$. This is the projected innovation matrix. 

Then: 
\begin{align*}
\operatorname{det}\left(\cakfcov_\idxdtime\right) & =\operatorname{det}\left(\cakfpcov_\idxdtime-\cakfpcov_\idxdtime \obsH_{k}^{\top} \cakfacts_k \cakfprojgram_{\idxdtime}^{-1} \cakfacts_k^{\top} \obsH_{k} \cakfpcov_\idxdtime\right) \\
& =\operatorname{det}\left(\sqrt{\cakfpcov_\idxdtime}\left[\mI-\sqrt{\cakfpcov_\idxdtime} \obsH_{k}^{\top} \cakfacts_k \cakfprojgram_{\idxdtime}^{-1} \cakfacts_k^{\top} \obsH \sqrt{\cakfpcov_\idxdtime}\right] \sqrt{\cakfpcov_\idxdtime}\right) \\
& =\operatorname{det}\left(\sqrt{\cakfpcov_\idxdtime}\right)^2 \operatorname{det}\left(\mI-\sqrt{\cakfpcov_\idxdtime} \obsH^{\top} \cakfacts_k \cakfprojgram_{\idxdtime}^{-1} \cakfacts_k^{\top} \obsH_k \sqrt{\cakfpcov_\idxdtime} \right)
\end{align*}

Define $\mL_{k} := \sqrt{\cakfpcov_\idxdtime} \obsH^{\top} \cakfacts_k$, such that $\mL_{k}^{\top} \mL_{k} = \cakfacts_k^{\top} \obsH \cakfpcov_\idxdtime \obsH^{\top} \cakfacts_k$
\begin{align*}
&=\operatorname{det}\left(\cakfpcov_\idxdtime\right) \operatorname{det}\left(\mI-\mL_{k} \cakfprojgram_{\idxdtime}^{-1} \mL_{k}^{\top}\right) \\
&=\operatorname{det}\left(\cakfpcov_\idxdtime\right) \operatorname{det}\left(\mI-\cakfprojgram_{\idxdtime}^{-1} \mL_{k}^{\top} \mL_{k}\right) \\
&=\operatorname{det}\left(\cakfpcov_\idxdtime\right) \operatorname{det}\left(\cakfprojgram_{\idxdtime}^{-1}\left(\cakfprojgram_{\idxdtime}-\mL_{k}^{\top} \mL_{k}\right)\right) \\
&=\operatorname{det}\left(\cakfpcov_\idxdtime\right) \operatorname{det}\left(\cakfprojgram_{\idxdtime}^{-1} \obsnoisecov_{k}\right) \\
&=\operatorname{det}\left(\cakfpcov_\idxdtime\right) \cdot \operatorname{det}\left(\obsnoisecov_{k}\right) \cdot \operatorname{det}\left(\cakfprojgram_{\idxdtime}\right)^{-1}
\end{align*}
Therefore, our original optimization problem becomes:
\begin{align*}
  \argmax_{\cakfacts_k} \mathrm{H}^{(k)} &= \argmax_{\cakfacts_k} \frac{1}{2} \log \left( \frac{\det \left(\cakfpcov_\idxdtime \right)}{\det \left( \cakfpcov_\idxdtime \right) \cdot \det \left(\obsnoisecov_{k}\right) \cdot \det \left( \cakfprojgram_{\idxdtime} \right)^{-1}} \right)\\
&= \argmax_{\cakfacts_k}  \frac{1}{2} \left[ \log \det \left( \cakfprojgram_{\idxdtime} \right) - \log (\det \left(\obsnoisecov_{k}\right)) \right]\\
&= \argmax_{\cakfacts_k}  \frac{1}{2} \left[ \log \det \left( \cakfprojgram_{\idxdtime} \right) \right]\\
&= \argmax_{\cakfacts_k}  \frac{1}{2} \left[\sum_{i=1}^m \log (d_i) \right]
\end{align*}

where $d_i > 0$ are the eigenvalues of the positive definite matrix $\cakfprojgram_{\idxdtime}$. 

The diagonalization of $\obsH_{k} \cakfpcov_\idxdtime \obsH_{k}^{\top}+\obsnoisecov_{k}= \kfgram_\idxdtime = \mU_m \mD \mU_m^T$ restricted to the top $m$ eigenvectors $\mU_m$ then yields the result by the Eckart-Young Theorem. 

$\cakfacts_k = \mU_m$ is a solution to this optimization problem.

\section{Smoothing Error Bound and Uncertainty Increase}

A core contribution of the CAKF \citep{Pfoertner2024ComputationAware} is the introduction of a smoothing error bound, which quantifies the increase in uncertainty from approximation during inference in the Kalman Smoother. 

CASSM preserves this interpretation during inference, though model misspecification is of course a higher-level source of error. 

\begin{figure*}[h]
  \centering
  \includegraphics[width=\textwidth]{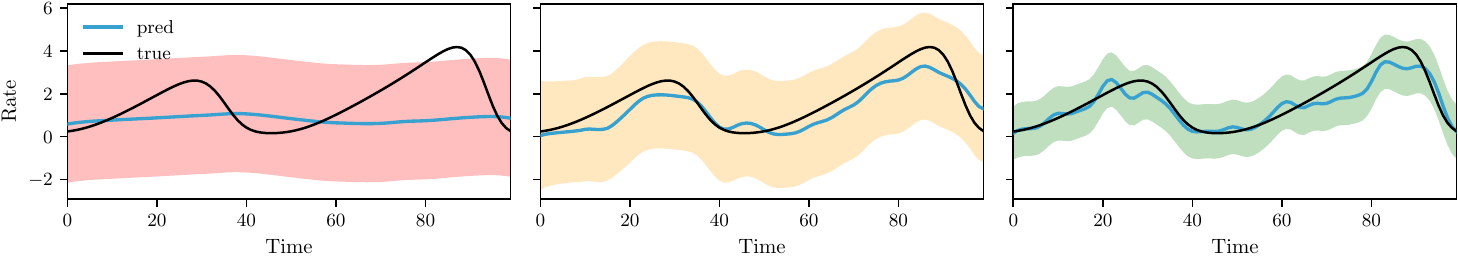}
  \caption{Moving from left to right, we see the effect of posterior approximation: (A) we project the Lorenz observations to a 2-dimensional approximate space, (B) we project to a 10-dimensional approximate space, (C) in the Kalman filter and smoother, we compute the full posterior of a 30-dimensional state space. Thus, the increase in uncertainty is \textit{relative} to the mean prediction, which is expected to be worse in a subspace that is insufficiently large to capure latent dynamics.}
  \label{fig:uncertainty-per-method}
\end{figure*}

In Figure \ref{fig:uncertainty-per-method}, we decompose the three cases of Figure \ref{fig:bands} to demonstrate that the increase in uncertainty is not spurious or automatic, but \textit{relative} to the mean prediction. 

\label{sec:bound}

\section{LTI-SDEs and the Matérn Kernel Formulation}
\label{sec:lti-sde}

The Matérn kernel is a widely used kernel in Gaussian Process regression. 

Here we give on overview of the integration between the Matérn kernel and the LTI-SDE formulation of the latent dynamics as in \citep{Sarkka2010gp}.

In the case of a continuous-time LTI-SDE, we have:
\begin{align*}
    d\mathbf{x} &= \mF \mathbf{x} \, dt + \mL \, d\boldsymbol{\beta},
\end{align*}
The drift matrix $\mF$ and dispersion matrix $\mL$ are constant, and $\boldsymbol{\beta}(t)$ has diffusion matrix $\mD$. The mean, $\mathbf{m}$, and covariance, $\mP$, are then given by:
\begin{align*}
    \frac{d\mathbf{m}}{dt} &= \mF \mathbf{m}, \\
    \frac{d\mP}{dt} &= \mF \mP + \mP \mF^T + \mL \mD  \mL^T.
\end{align*}
whose solution, in the case of our discretized dynamics model:
\begin{align*}
    \mathbf{x}(t_{k+1}) &= \mA_k \mathbf{x}(t_k) + \mathbf{q}_k, \quad \mathbf{q}_k \sim \mathcal{N}(0, \gmpnoisecov_k)
\end{align*}
where \( \Delta t_k = t_{k+1} - t_k \), is given by:
\begin{align*}
    \mA_k &= \exp(\mF \Delta t_k), \\
    \gmpnoisecov_k &= \int_0^{\Delta t_k} \exp(\mF (\Delta t_k - \tau)) \mL \mD \mL^T \exp(\mF (\Delta t_k - \tau))^T d\tau.
\end{align*}

For stable drift matrices, solving the Lyapunov differential equation $\frac{d\mP}{dt}$ yields a steady state covariance $\gmpcov_\infty$.

Now, in general, for a space-time separable Gauss-Markov Process, the covariance factorizes along time and space, or:
\begin{align*}
  \gmpcov\left[ (t_1, \mathbf{x}_1), (t_2, \mathbf{x}_2)\right] = \gmpcov^{(\mathbf{x})}(\mathbf{x}_1, \mathbf{x}_2) \cdot \gmpcov^{(t)} (t_1, t_2) 
\end{align*}
This observation is the key insight that allows Gaussian Process regressions to be solved in linear time under certain classes of kernel functions $\gmpcov^{(\cdot)}$. Under the Matern$(\nu = p + \frac{1}{2})$ class of kernel, the state-space representation is indeed exact \citep{Sarkka2010gp}. 

Under separability, the stationary covariance admits a Kronecker factorization:
\begin{align*}
  \gmpcov_{\infty} = \gmpcov^{(\mathbf{x})}(\mathbf{x}_1, \mathbf{x}_2) \otimes \gmpcov_{\infty}^{(t)} (t_1, t_2)
\end{align*}
Similarly, the drift matrix admits a Kronecker factorization $\mF = \mI_{d \times d} \otimes \mF^{(t)}$ as the neurons only change in time and not space. 

Thus, the matrix exponential of $\mF$, which is otherwise a prohibitively expensive calculation, only scales unfavorably (cubically) with the respect to the smoothness of the time kernel $p_t$ which reasonably is a small integer value below ten.

Indeed, in the case where the time dynamics can be described by a Matern kernel, $\mF^{(t)}$ and $\gmpcov_{\infty}^{(t)}$ are given analytically for $p_t=0,1,2$, and there is no need to solve the Lyapunov differential equation.

In any case, once $\gmpcov_\infty$ is obtained, this allows the process noise covariance, for any arbitrary timepoint, to be computed as:
\begin{align*}
  \gmpnoisecov_k &= \gmpcov_\infty - \mA_k \gmpcov_\infty \mA_k^T
\end{align*}

\section{Experiments}
\label{sec:experimental-details}

\subsection{Hardware}
\label{sec: hardware}

All experiments except the Zebrafish experiment were run on a single NVIDIA GeForce RTX 4090 with 24GB of VRAM and an 
Intel Core i9-14900KF with 64GB of DRAM. 

The Zebrafish experiment was run on a single NVIDIA A100 with 80GB of VRAM and an Intel Xeon Platinum CPU with 512 GB of DRAM.

\subsection{Lorenz Experiments}

\subsubsection{Computational Budget vs. Time to Convergence}
\label{sec:computational-budget}

In Figure \ref{fig:scaling}, we ran the algorithms for a fixed computational budget, there defined as a fixed number of passes through the data. 

The computation-aware regime mainly considers the problem of limited compute, which is especially relevant for GPFA and the Kalman Filter, as running these to convergence at higher neural scales would require several days per run and thus several weeks total for a sufficient number of seeds. 

Because of our problem statement, the practicality considerations of running these algorithms repeatedly, and our preliminary experiments of running algorithms to convergence, we chose to report all algorithms for a fixed number of passes through the data. 

\begin{figure}
  \centering
  \includegraphics[width=0.4\textwidth]{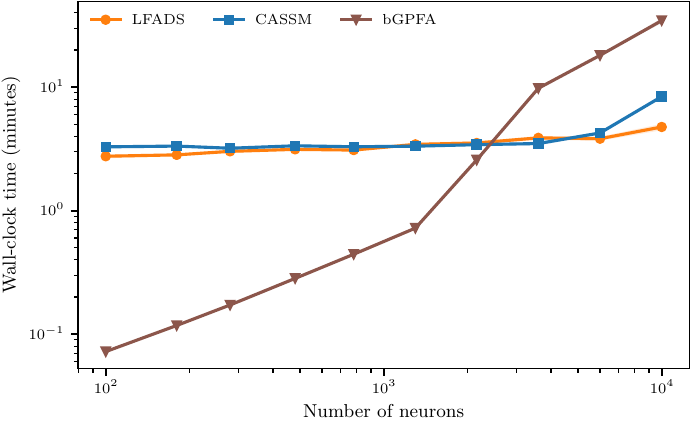}
  \caption{Here we report time to convergence instead of a fixed computational budget. CASSM becomes slightly slower relative to LFADS but still does not begin to enter its linear scaling regime until around $10^4$ neurons, whereas bGPFA still begins its quasilinear regime much earlier.}
  \label{fig:scaling-convergence}
 \end{figure}

However, an equally important choice of reporting is the scaling with respect to time to convergence, which we report in Figure \ref{fig:scaling-convergence}. 

Here, we show the three feasible methods at higher neural scales, and see that the ordering does not change. 

CASSM takes slightly longer relative to LFADS to converge at these higher neural scales, but the trends in both perspectives mostly hold for the feasible methods. 

We note that the time to convergence for GPFA and the Kalman Filter would yield sharper curves, but these methods are not the focus of the paper. 

\subsubsection{Hyperparameters}
\label{sec:lorenz-hyperparameters}
While LFADS has an array of hyperparameters that are chosen via the original paper \citep{Sussillo2016lfads} in that version of the Lorenz experiment, the major hyperparameters for the Lorenz experiments are as follows:

\begin{table}[h]
  \centering
  \normalsize
  \renewcommand{\arraystretch}{1.05}
  \setlength{\tabcolsep}{6pt}
  \begin{tabular*}{\columnwidth}{@{\extracolsep{\fill}} l ll r @{}}
      \toprule
      \multicolumn{4}{c}{Training Hyperparameters} \\
      \midrule
      Method & Optimizer & Learning Rate \\
      \midrule
      LFADS \citep{Pandarinath2018LFADS}        & Adam      & $0.100$ \\
      GPFA        \citep{Yu2008GPFA}            &  LBFGS     & $0.050$ \\
      bGPFA \citep{Jensen2021Scalable}                         & Adam      & $0.010$ \\
      Kalman                 &  Adam     & 0.001 \\
      CASSM                        & Adam      & $0.005$ \\
      \bottomrule
  \end{tabular*}
  \caption{Optimizers and learning rates used for each method.}
  \label{tab:method-optim-lr}
\end{table}

These were obtained by grid search over the following learning rates for each method:

\[\{0.0001, 0.0005, 0.001, 0.005, 0.01, 0.05, 0.1, 0.5, 1.0\} \]

The non-LFADS methods are kernel-based and use a Matérn$\left( \frac{1}{2} \right)$ time kernel. 

Note that the Laplacian kernel used in GPFA is equivalent to a Matérn$\left( \frac{1}{2} \right)$ kernel. Via space-time separability, CASSM also has a spatial Matérn$\left( \frac{3}{2} \right)$ kernel. 

This spatial kernel has an outputscale $\sigma^2$ and one lengthscale $\ell_j$ per spatial input dimension, which is always $3$ regardless of dataset due to the spatial location of each individual neuron.

All of the methods are given $3$ latent factors, which is standard for the Lorenz experiment. 

We note that bGPFA also has an additional hyperparameter, the number of Monte Carlo samples. 

We pushed this up to the highest number that we could afford computationally, which was $10$ samples at the larger scales (past $10^3$ neurons) and $100$ samples at the smaller scales. 

We note that setting this too high causes out-of-memory errors, and setting it too low causes higher variance estimates. 

\subsection{Neural Latents Benchmark Experiments}

The Neural Latents Benchmark \citep{PeiYe2021NeuralLatents} is a collection of datasets that are designed to evaluate the performance of latent variable models on real neural data. 

The datasets consist of neural recordings from various brain regions and species, along with corresponding behavioral data. 

\subsubsection{Evaluation Metrics}
\label{sec:nlb-metrics}

The Neural Latents Benchmark (NLB) presents several evaluation metrics which assess different aspects of model performance on neural population data.

\paragraph{Co-smoothing (co-bps)}
The co-smoothing metric evaluates how well a model predicts the firing activity of held-out neurons given the activity of held-in neurons. 

Concretely, models are trained on a subset of neurons and must predict firing rates for the remaining neurons. 

Performance is measured as the log-likelihood improvement (in bits per spike) relative to a baseline model that predicts the trial-averaged firing rate. 

This metric captures how well the inferred latent dynamics generalize across neurons and reflects the quality of the learned population-level representation \cite{PeiYe2021NeuralLatents}. 

\paragraph{Velocity $R^2$ (vel $R^2$)}
The velocity $R^2$ metric measures how well the inferred neural representations can decode behavior, like hand velocity in motor tasks. 

A linear decoder is trained to map inferred firing rates (or latent features) to kinematic variables, and the coefficient of determination ($R^2$) is computed between predicted and true velocities. 

This metric evaluates whether the model preserves behaviorally relevant information in the neural population activity and serves as a proxy for downstream decoding performance.

\paragraph{Peristimulus Time Histogram (psth $R^2$)}
The peristimulus $R^2$ metric evaluates how well the model captures trial-averaged neural responses. 

For each neuron, the peristimulus time histogram (PSTH) is computed by averaging firing rates across trials aligned to a stimulus or behavioral event. 

The coefficient of determination ($R^2$) is then computed between the predicted and true PSTHs. 

This metric emphasizes the model's ability to capture consistent, stimulus-locked structure in neural activity while averaging out trial-to-trial variability.

\subsubsection{Hyperparameters}

We repeated our grid search for learning rates on the NLB datasets, except in this case the LFADS and GPFA models were tuned by the original authors of the NLB, and we used their recommended learning rates for those methods, a process described in that original paper. 

The other methods were tuned by us via grid search over the same learning rates as in the Lorenz experiment. The results are summarized in Table \ref{tab:method-optim-lr-neural-latents}.

\begin{table}[t]
  \centering
  \normalsize
  \renewcommand{\arraystretch}{1.05}
  \setlength{\tabcolsep}{6pt}
  \begin{tabular*}{\columnwidth}{@{\extracolsep{\fill}} l l r @{}}
      \toprule
      \multicolumn{3}{c}{Training Hyperparameters by Dataset} \\
      \midrule
      
      \multicolumn{3}{c}{\textbf{Dataset 1: MC Maze}} \\
      \midrule
      Method & Optimizer & Learning Rate \\
      \midrule
      bGPFA \citep{Jensen2021Scalable}   & Adam  & 0.010 \\
      Kalman                              & Adam  & 0.001 \\
      CASSM                               & Adam  & 0.005 \\
      
      \midrule
      \multicolumn{3}{c}{\textbf{Dataset 2: MC RTT}} \\
      \midrule
      Method & Optimizer & Learning Rate \\
      \midrule
      bGPFA \citep{Jensen2021Scalable}   & Adam  & 0.050 \\
      Kalman                              & Adam  & 0.01 \\
      CASSM                               & Adam  & 0.001 \\
      
      \midrule
      \multicolumn{3}{c}{\textbf{Dataset 3: DMFC RSG}} \\
      \midrule
      Method & Optimizer & Learning Rate \\
      \midrule
      bGPFA \citep{Jensen2021Scalable}   & Adam  & 0.050 \\
      Kalman                              & Adam  & 0.001 \\
      CASSM                               & Adam  & 0.010 \\
      
      \midrule
      \multicolumn{3}{c}{\textbf{Dataset 4: Area 2 Bump}} \\
      \midrule
      Method & Optimizer & Learning Rate \\
      \midrule
      bGPFA \citep{Jensen2021Scalable}   & Adam  & 0.010 \\
      Kalman                              & Adam  & 0.005 \\
      CASSM                               & Adam  & 0.005 \\
      
      \bottomrule
  \end{tabular*}
  \caption{Optimizers and learning rates used for each method across the Neural Latents benchmark datasets.}
  \label{tab:method-optim-lr-neural-latents}
\end{table}

The models were saved at the epoch of convergence, where convergence was defined as the epoch at which the validation loss improvement was less than $10^{-5}$ for $5$ epochs.

Unlike the previous Lorenz experiments, the number of latent factors were not fixed across methods where a ground truth existed. 

Instead, we tuned the number of latent factors for each method by grid search over $5$ to $50$ latent factors with a step size of $5$. 

The best performing number of latent factors for each method and dataset was chosen by the lowest validation loss, which was consistently around $10$ or $15$ factors per dataset and method, likely reflecting some information about the intrinsic dimensionality of each dataset. 

We note that bGPFA learns an appropriate latent dimensionality using automatic relevance determination, starting with a max dimension and letting the model turn off irrelevant latents. 

\subsection{Zebrafish Experiments}

The Zebrafish dataset evaluation is similar to the Neural Latents Benchmark, except that the LFADS model was not tuned by the original authors of the dataset, and we had to perform our own grid search for learning rates for that method as well. 

In addition, the evaluation metrics are purely fit-based as the data types differ: we are not predicting firing rates or spikes, but fluorescence activity. 

Due to that structure, this data is actually more suitable for the Gaussian likelihood, and we do not perform any normalization or preprocessing (smoothing, etc.) for CASSM in this case. 

The raw data is fed directly to the model. The results are summarized in Table \ref{tab:method-optim-lr-zebra}. 

We held out $25\%$ randomly-selected test neurons for the zebrafish dataset, which was not used for training or validation, and we report test performance on that held-out data. 

In addition, we initially subsampled a number of neurons, $30,000$ from approximately $95,000$ to keep our analysis consistent with our Lorenz experiments and to speed up inference given that each method was fit on a grid. 

Even in this regime, we could not choose any number of latent factors or Monte Carlo samples for bGPFA without causing out-of-memory errors. 

We note that the original bGPFA paper only goes up to a maximum of $200$ neurons, so this is not particularly surprising. 

The models were saved at the epoch of convergence, where convergence was again defined as the epoch at which the validation loss improvement was less than $10^{-5}$ for $5$ epochs.
\begin{table}[h]
  \centering
  \normalsize
  \renewcommand{\arraystretch}{1.05}
  \setlength{\tabcolsep}{6pt}
  \begin{tabular*}{\columnwidth}{@{\extracolsep{\fill}} l ll r @{}}
      \toprule
      \multicolumn{4}{c}{Training Hyperparameters} \\
      \midrule
      Method & Optimizer & Learning Rate \\
      \midrule
      LFADS \citep{Pandarinath2018LFADS}        & Adam      & $0.100$ \\
      bGPFA \citep{Jensen2021Scalable}                         & Adam      & $--$ \\
      CASSM                        & Adam      & $0.005$ \\
      \bottomrule
  \end{tabular*}
  \caption{Optimizers and learning rates used for each method.}
  \label{tab:method-optim-lr-zebra}
\end{table}

For the number of latent factors, we performed grid search over $10$ to $100$ latent factors with a step size of $10$, and the best performing number of latent factors for each method was chosen by the lowest validation loss, which was $20$ latent factors for the CASSM variants and $30$ latent factors for LFADS. 
\stopcontents[sections]

%% file: references.bib
@String{NeurIPS     = {Advances in Neural Information Processing Systems (NeurIPS)}}

@String{ICLR        = {International Conference on Learning Representations (ICLR)}}

@String{ARXIV       = {arXiv}}

@String{MIT         = {MIT Press}}

@String{Elsevier    = {Elsevier}}

@String{CUP         = {Cambridge University Press}}

@article{PeiYe2021NeuralLatents,
  title     = {Neural Latents Benchmark '21: Evaluating latent variable models of neural population activity},
  author    = {Felix Pei and Joel Ye and David M. Zoltowski and Anqi Wu and Raeed H. Chowdhury and Hansem Sohn and Joseph E. O’Doherty and Krishna V. Shenoy and Matthew T. Kaufman and Mark Churchland and Mehrdad Jazayeri and Lee E. Miller and Jonathan Pillow and Il Memming Park and Eva L. Dyer and Chethan Pandarinath},
  booktitle = {Advances in Neural Information Processing Systems (NeurIPS), Track on Datasets and Benchmarks},
  year      = {2021},
  url       = {https://arxiv.org/abs/2109.04463}
}

@article{zhang2024differentiablesvdbasedmoorepenrose,
  title  = {Differentiable SVD based on Moore-Penrose Pseudoinverse for Inverse Imaging Problems},
  author = {Yinghao Zhang and Yue Hu},
  year   = {2024},
  eprint = {2411.14141}
}

@article{Sarkka2010gp,
  title       = {KALMAN FILTERING AND SMOOTHING SOLUTIONS TO TEMPORAL GAUSSIAN PROCESS REGRESSION MODELS},
  author      = {Sarkka, Simo},
  institution = {Aalto University},
  year        = {2010},
  url         = {https://users.aalto.fi/~ssarkka/pub/gp-ts-kfrts.pdf}
}

@article{Fu2022hungry,
  title   = {Hungry Hungry Hippos: Towards Language Modeling with State Space Models},
  author  = {Fu, Daniel Y. and Dao, Tri and Saab, Khaled K. and Thomas, Armin W. and Rudra, Atri and R{\'e}, Christopher},
  journal = {arXiv preprint arXiv:2212.14052},
  year    = {2022},
  note    = {ICLR 2023},
  doi     = {10.48550/arXiv.2212.14052},
  url     = {https://arxiv.org/abs/2212.14052}
}

@article{Wang2018batched,
  title         = {Batched Large-scale Bayesian Optimization in High-dimensional Spaces},
  author        = {Zi Wang and Clement Gehring and Pushmeet Kohli and Stefanie Jegelka},
  year          = {2018},
  eprint        = {1706.01445},
  archiveprefix = {arXiv},
  primaryclass  = {stat.ML},
  url           = {https://arxiv.org/abs/1706.01445}
}

@article{Gallego2017neural,
  title     = {Neural manifolds for the control of movement},
  author    = {Gallego, Juan A and Perich, Matthew G and Miller, Lee E and Solla, Sara A},
  journal   = {Neuron},
  volume    = {94},
  number    = {5},
  pages     = {978--984},
  year      = {2017},
  publisher = {Elsevier}
}

@article{Pfoertner2024ComputationAware,
  author    = {Pf\"ortner, Marvin and Wenger, Jonathan and Cockayne, Jon and Hennig, Philipp},
  title     = {Computation-Aware {K}alman Filtering and Smoothing},
  year      = {2024},
  publisher = {arXiv},
  doi       = {10.48550/arxiv.2405.08971},
  url       = {https://arxiv.org/abs/2405.08971}
}

@article{Jensen2021Scalable,
  author       = {Jensen, Kristopher T. and Kao, Ta-Chu and Stone, Jasmine T. and Hennequin, Guillaume},
  title        = {Scalable {Bayesian} {GPFA} with automatic relevance determination and discrete noise models},
  elocation-id = {2021.06.03.446788},
  year         = {2021},
  doi          = {10.1101/2021.06.03.446788},
  publisher    = {Cold Spring Harbor Laboratory},
  url          = {https://www.biorxiv.org/content/early/2021/06/03/2021.06.03.446788},
  eprint       = {https://www.biorxiv.org/content/early/2021/06/03/2021.06.03.446788.full.pdf},
  journal      = {bioRxiv}
}

@article{duncker2018temporal,
  title   = {Temporal alignment and latent Gaussian process factor inference in population spike trains},
  author  = {Duncker, Lea and Sahani, Maneesh},
  journal = NeurIPS,
  volume  = {31},
  year    = {2018}
}

@article{Karpowicz2022BCI,
  author       = {Karpowicz, Brianna M. and Ali, Yahia H. and Wimalasena, Lahiru N. and Sedler, Andrew R. and Keshtkaran, Mohammad Reza and Bodkin, Kevin and Ma, Xuan and Miller, Lee E. and Pandarinath, Chethan},
  title        = {Stabilizing brain-computer interfaces through alignment of latent dynamics},
  elocation-id = {2022.04.06.487388},
  year         = {2022},
  doi          = {10.1101/2022.04.06.487388},
  publisher    = {Cold Spring Harbor Laboratory},
  eprint       = {https://www.biorxiv.org/content/early/2022/04/08/2022.04.06.487388.full.pdf},
  journal      = {bioRxiv}
}

@article{Zhu2021Deepinference,
  title         = {Deep inference of latent dynamics with spatio-temporal super-resolution using selective backpropagation through time},
  author        = {Feng Zhu and Andrew R. Sedler and Harrison A. Grier and Nauman Ahad and Mark A. Davenport and Matthew T. Kaufman and Andrea Giovannucci and Chethan Pandarinath},
  year          = {2021},
  eprint        = {2111.00070},
  archiveprefix = {arXiv},
  primaryclass  = {cs.LG},
  url           = {https://arxiv.org/abs/2111.00070},
  publisher     = {{NeurIPS}}
}

@article{Sussillo2016lfads,
  title         = {LFADS - Latent Factor Analysis via Dynamical Systems},
  author        = {David Sussillo and Rafal Jozefowicz and L. F. Abbott and Chethan Pandarinath},
  year          = {2016},
  eprint        = {1608.06315},
  archiveprefix = {arXiv},
  primaryclass  = {cs.LG},
  url           = {https://arxiv.org/abs/1608.06315}
}

@article{Pnevmatikakis2014FastKalman,
  author  = {Pnevmatikakis, Eftychios A. and Rahnama Rad, Kamiar and Huggins, Justin and Paninski, Liam},
  title   = {Fast Kalman Filtering and Forward–Backward Smoothing via a Low-Rank Perturbative Approach},
  journal = {Journal of Computational and Graphical Statistics},
  volume  = {22},
  number  = {3},
  pages   = {671–690},
  year    = {2014},
  doi     = {10.1080/10618600.2012.760461}
}

@article{Paninski2010FastKalmanDendritic,
  author  = {Paninski, Liam},
  title   = {Fast Kalman filtering on quasilinear dendritic trees},
  journal = {Journal of Computational Neuroscience},
  volume  = {28},
  number  = {2},
  pages   = {211–228},
  year    = {2010},
  doi     = {10.1007/s10827-009-0200-4}
}

@inproceedings{Macke2011Empirical,
  title     = {Empirical models of spiking in neural populations},
  author    = {Jakob H. Macke and Lars B\"using and John P. Cunningham and Byron M. Yu and Krishna V. Shenoy and Maneesh Sahani},
  booktitle = {Advances in Neural Information Processing Systems 24 (NIPS 2011)},
  pages     = {1350--1358},
  year      = {2011},
  url       = {https://papers.nips.cc/paper/2011/file/7143d7fbadfa4693b9eec507d9d37443-Paper.pdf}
}

@article{Wang2021RobustDifferentiableSVD,
  title         = {Robust Differentiable SVD},
  author        = {Wei Wang and Zheng Dang and Yinlin Hu and Pascal Fua and Mathieu Salzmann},
  journal       = {arXiv preprint arXiv:2104.03821},
  year          = {2021},
  eprint        = {2104.03821},
  archiveprefix = {arXiv},
  primaryclass  = {cs.CV},
  doi           = {10.1109/TPAMI.2021.3072422}
}

@article{Bonnabel2012GeometryLowRank,
  author  = {Bonnabel, Silv{\`e}re and Sepulchre, Rodolphe},
  title   = {The geometry of low-rank Kalman filters},
  journal = {arXiv preprint},
  eprint  = {1203.4049},
  year    = {2012},
  note    = {Revised version: 2013},
  url     = {https://arxiv.org/abs/1203.4049}
}

@article{Pandarinath2018LFADS,
  title   = {Inferring single-trial neural population dynamics using sequential auto-encoders},
  author  = {Pandarinath, Chethan and O'Shea, Daniel J. and Collins, Jasmine and Jozefowicz, Rafal and Stavisky, Sergey D. and Kao, Jonathan C. and Trautmann, Eric M. and Kaufman, Matthew T. and Ryu, Stephen I. and Hochberg, Leigh R. and Henderson, Jaimie M. and Shenoy, Krishna V. and Abbott, L. F. and Sussillo, David},
  journal = {Nature Methods},
  year    = {2018},
  volume  = {15},
  number  = {10},
  pages   = {805--815},
  month   = {oct},
  doi     = {10.1038/s41592-018-0109-9},
  url     = {https://www.nature.com/articles/s41592-018-0109-9}
}

@article{Yu2008GPFA,
  author    = {Yu, Byron M and Cunningham, John P and Santhanam, Gopal and Ryu, Stephen and Shenoy, Krishna V and Sahani, Maneesh},
  booktitle = NeurIPS,
  editor    = {D. Koller and D. Schuurmans and Y. Bengio and L. Bottou},
  publisher = {Curran Associates, Inc.},
  title     = {Gaussian-process factor analysis for low-dimensional single-trial analysis of neural population activity},
  url       = {https://proceedings.neurips.cc/paper_files/paper/2008/file/ad972f10e0800b49d76fed33a21f6698-Paper.pdf},
  volume    = {21},
  year      = {2008}
}

@article{Zhao2017VLGP,
  author    = {Yuan Zhao and Il Memming Park},
  title     = {Variational Latent Gaussian Process for Recovering Single-Trial Dynamics from Population Spike Trains},
  journal   = {Neural Computation},
  year      = {2017},
  volume    = {29},
  number    = {5},
  pages     = {1293--1316},
  month     = {may},
  doi       = {10.1162/neco_a_00953},
  publisher = {{MIT} Press - Journals}
}

@article{Steinmetz2019Distributed,
  author  = {Steinmetz, Nicholas and Zatka-Haas, Peter and Carandini, Matteo and Harris, Kenneth},
  journal = {Nature},
  title   = {Distributed coding of choice, action, and engagement across the mouse brain},
  year    = {2019}
}

@article{Siegle2021Spiking,
  author  = {Siegle, J.H. and Jia, Xiaoxuan and Durand, Severine},
  journal = {Nature},
  title   = {Survey of spiking in the mouse visual system reveals functional hierarchy},
  year    = {2021}
}

@inproceedings{Wenger2022PosteriorComputational,
  author    = {Jonathan Wenger and Geoff Pleiss and Marvin Pf{\"o}rtner and Philipp Hennig and John P. Cunningham},
  booktitle = NeurIPS,
  title     = {Posterior and Computational Uncertainty in {G}aussian processes},
  year      = {2022}
}

@book{Sarkka2023BayesianFiltering,
  edition   = {2nd},
  title     = {Bayesian {Filtering} and {Smoothing}},
  volume    = {17},
  isbn      = {978-1-108-91230-3},
  language  = {en},
  publisher = CUP,
  author    = {Särkkä, Simo and Svensson, Lennart},
  year      = {2023}
}

@article{Keshtkaran2022AutoLFADS,
  title   = {A large-scale neural network training framework for generalized estimation of single-trial population dynamics},
  author  = {Keshtkaran, Mohammad Reza and Sedler, Andrew R. and Chowdhury, Raeed H. and Tandon, Raghav and Basrai, Diya and Nguyen, Sarah L. and Sohn, Hansem and Jazayeri, Mehrdad and Miller, Lee E. and Pandarinath, Chethan},
  journal = {Nature Methods},
  year    = {2022},
  volume  = {19},
  number  = {12},
  pages   = {1572--1577},
  doi     = {10.1038/s41592-022-01675-0},
  pmid    = {36443486},
  pmcid   = {PMC9825111}
}

@article{Wenger2024computationawaregp,
  title     = {Computation-Aware Gaussian Processes: Model Selection and Linear-Time Inference},
  author    = {Wenger, Jonathan and Wu, Kaiwen and Hennig, Philipp and Gardner, Jacob R. and Pleiss, Geoff and Cunningham, John P.},
  booktitle = {Advances in Neural Information Processing Systems (NeurIPS)},
  year      = {2024}
}

@article{hoffmann2022training,
  title         = {Training Compute-Optimal Large Language Models},
  author        = {Hoffmann, Jordan and Borgeaud, Sebastian and Mensch, Arthur and Buchatskaya, Elena and Cai, Trevor and Rutherford, Eliza and de Las Casas, Diego and Hendricks, Lisa Anne and Welbl, Johannes and Clark, Aidan and Hennigan, Tom and Noland, Eric and Millican, Katie and van den Driessche, George and Damoc, Bogdan and Guy, Aurelia and Osindero, Simon and Simonyan, Karen and Elsen, Erich and Rae, Jack W. and Vinyals, Oriol and Sifre, Laurent},
  journal       = {arXiv preprint},
  volume        = {2203.15556},
  year          = {2022},
  archiveprefix = {arXiv},
  primaryclass  = {cs.CL},
  url           = {https://arxiv.org/abs/2203.15556}
}

@article{Gauthaman2025scaleFree,
  title   = {Universal scale-free representations in human visual cortex},
  author  = {Gauthaman, Raj Magesh and M{\'e}nard, Brice and Bonner, Michael F.},
  journal = {PLOS Computational Biology},
  year    = {2025},
  volume  = {21},
  number  = {11},
  pages   = {e1013714},
  doi     = {10.1371/journal.pcbi.1013714},
  url     = {https://doi.org/10.1371/journal.pcbi.1013714}
}

@article{Chen2018Zebra,
  author  = {Chen, X. and Mu, Y. and Hu, Y. and Kuan, A. T. and Nikitchenko, M. and Randlett, O. and Chen, A. B. and Gavornik, J. P. and Sompolinsky, H. and Engert, F. and Ahrens, M. B.},
  title   = {Brain-wide Organization of Neuronal Activity and Convergent Sensorimotor Transformations in Larval Zebrafish},
  journal = {Neuron},
  volume  = {100},
  number  = {4},
  pages   = {876--890.e5},
  year    = {2018},
  doi     = {10.1016/j.neuron.2018.09.064}
}

@misc{Gu2023MambaLinearTime,
  title      = {Mamba: {Linear}-{Time} {Sequence} {Modeling} with {Selective} {State} {Spaces}},
  shorttitle = {Mamba},
  url        = {http://arxiv.org/abs/2312.00752},
  doi        = {10.48550/arXiv.2312.00752},
  publisher  = {arXiv},
  author     = {Gu, Albert and Dao, Tri},
  year       = {2023}
}
